\documentclass[10pt,twocolumn]{article}

\usepackage[margin=1in,columnsep=0.25in]{geometry}
\usepackage{amsmath,amssymb,amsfonts}
\usepackage{graphicx}
\usepackage{booktabs}
\usepackage{multirow}
\usepackage{array}
\usepackage{tabularx}
\usepackage{siunitx}
\usepackage[colorlinks=true,linkcolor=blue,citecolor=blue,urlcolor=blue]{hyperref}
\usepackage{url}
\usepackage{tikz}
\usepackage{xcolor}
\usepackage{cite}
\usepackage{caption}
\usepackage{subcaption}
\usepackage{enumitem}

\usepackage{pgfplots}
\pgfplotsset{compat=1.18}
\usetikzlibrary{positioning,arrows.meta,calc,fit,shapes.geometric,%
backgrounds,decorations.pathreplacing,patterns}
\usepackage{tcolorbox}
\tcbuselibrary{skins,breakable}

\definecolor{paperblue}{RGB}{46,89,167}    % data / inputs
\definecolor{papergreen}{RGB}{34,120,74}   % backbone / model
\definecolor{paperred}{RGB}{178,34,34}     % physics priors
\definecolor{paperviolet}{RGB}{104,60,157} % retrieval / memory
\definecolor{paperorange}{RGB}{204,102,0}  % predictions / outputs

\IfFileExists{stix2.sty}{\usepackage{stix2}}{} % fall back to the default serif if unavailable
\usepackage[expansion=false]{microtype}

\usepackage{titlesec}
\titleformat{\section}{\normalfont\bfseries\large}{\thesection.}{0.5em}{}
\titleformat{\subsection}{\normalfont\bfseries\normalsize}{\thesubsection}{0.5em}{}
\titleformat{\subsubsection}{\normalfont\itshape\normalsize}{\thesubsubsection}{0.5em}{}

\usepackage{fancyhdr}
\title{%
  \vspace{-1em}
  \Large\textbf{Modular Deep Learning Mechanisms for Auditable Next-Day Wildfire Spread Prediction}%
  \vspace{0.3em}
}

\author{%
  \begin{tabular}{c}
    Miguel Esparza\textsuperscript{1,*} \quad
    Aydin Ayanzadeh\textsuperscript{2} \quad
    Ahmad Mousavi\textsuperscript{3} \quad
    Ali Mostafavi\textsuperscript{1}\\[0.15em]
    {\small *Corresponding author: \texttt{mte1224@tamu.edu}}\\[0.6em]
    \small\textsuperscript{1}Urban Resilience.AI Lab, Zachry Department of Civil
      and Environmental Engineering,\\
    \small Texas A\&M University, College Station, TX 77843, USA\\[0.2em]
    \small\textsuperscript{2}Department of Information Systems, University of
      Maryland, Baltimore County, Baltimore, MD 21250, USA\\[0.2em]
    \small\textsuperscript{3}Department of Mathematics and Statistics, American
      University, Washington, D.C. 20016, USA\\[0.4em]
    \small\texttt{aydina1@umbc.edu} \quad
    \texttt{mousavi@american.edu} \quad
    \texttt{amostafavi@civil.tamu.edu}
  \end{tabular}
}

\date{}

\begin{document}
\maketitle
\thispagestyle{fancy}

% ============================================================
\begin{abstract}

Next-day wildfire prediction requires models whose forecasts can be evaluated alongside the assumptions and historical evidence used in their computation. Although deep learning can learn spatial patterns from remote-sensing data, predictive performance alone does not establish physical fidelity or operational trustworthiness. This study investigates three modular augmentations for next-day active-fire prediction: wind- and slope-conditioned attention biases, physics-feature retrieval-augmented output correction, and fire conditioned dual-stream gating. The attention biases expose prescribed directional preferences, while the retrieval module selects historical tiles using a nine-dimensional environmental and fire-state descriptor and applies a learned correction to a frozen model's logits. The modules are evaluated across five backbones on the Next Day Wildfire Spread benchmark, using staged ablations, directional audits, retrieval perturbations, calibration measures, and computational comparisons. The three-seed mean F1 score and area under the precision--recall curve (AUC-PR) of a SwinUNETR model with all three augmentations are 0.4216 and 0.3673. Then, a mixed ensemble (two augmented architectures and one non-augmented architecture) model  achieves 0.4292 and 0.3790. Benefits vary across architectures, and retrieval-related improvements in AUC-PR do not consistently translate into higher F1. The constructed wind bias aligns closely with input wind, but its alignment with observed next-day fire displacement is much weaker, distinguishing prior inspectability from predictive physical fidelity. The study contributes a framework for exposing and evaluating selected domain-informed components within wildfire prediction models. Together, the results presented show that predictive performance, operational trustworthiness, and computational practicality need not be competing objectives.

\end{abstract}


\vspace{0.5em}
\hrule
\vspace{0.5em}

% ============================================================
\section{Introduction}
\label{sec:intro}
% ============================================================
During an active wildfire, the next day's fire location is a crucial component for planning. Forecasts can inform assessments of potential exposure, resource needs, and areas requiring closer observation. However, the forecast's value depends on both predictive skill and an understanding of their limitations. Satellite-based models provide a way to estimate next-day fire activity over large regions from prior fire observations and environmental conditions~\cite{huot2022}. The task remains difficult because fire detections are sparse, environmental inputs represent processes at different spatial and temporal scales, and observed changes in active fire do not capture every aspect of the evolving fire front. A useful forecasting approach must therefore be evaluated for its errors as well as for the evidence it exposes to the people interpreting its outputs. 

Recent deep learning approaches improve representations of spatial context and interactions among environmental variables~\cite{li2024asufm,chen2024autost,han2026}, yet prediction maps alone reveal little about the specific priors or precedents involved in a forecast. Domain-informed computation offers opportunities to expose these elements directly, but the resulting inspection capability should be distinguished from proof of physically correct behavior. 
The research explores the following: (1) If directional environmental cues can be made explicit inside the attention computation; (2) If historical fire observations provide traceable corrections; (3) Determine if the proposed additions remain useful across different deep learning architectures. These are addressed through wind- and slope-conditioned attention, physics-feature retrieval correction, and fire-conditioned gating. Evaluating the modules across five backbones allows their predictive effects and inspection properties to be examined together without assuming that one architecture's response will generalize to every other design.

In this study, next-day predictions refers to forecasting fire activity at the daily level while an incident is ongoing. Moreover, the prediction is framed as a pixel-level binary segmentation task on multimodal remote-sensing inputs that describe topography, vegetation, and meteorology. This model predicts which pixels will be burning at the next time step, $t{+}1$. Physics-based simulators represent this evolution mechanistically but require dense, well-calibrated fuel and weather inputs that are rarely available in real time. Deep learning models instead learn spread relationships from historical remote-sensing data and produce forecasts at the speed and spatial coverage an active incident demands. For example, Li and Rad (2024) coupled spatial attention with focal modulation inside a Swin transformer U-Net to forecast next-day spread across North America~\cite{li2024asufm}, and Chen et al.\ (2024) paired a three-dimensional convolutional network with a transformer in an encoder--decoder design~\cite{chen2024autost}. Two challenges accompany this progress. The first is transparency. Rudin (2019) argued that black-box models should have minimal use in high-stakes decisions that affect human lives, as they do not explain their predictions in a way that decision makers can understand~\cite{rudin2019}. This has also been documented across wildfire literature~\cite{becker2026assessing}. To mitigate this black box, researchers have used explainable AI (xAI) methods, such as SHAP or integrated gradients, to assess a model's alignment with fire science~\cite{becker2026assessing, vu2025black}. These methods produce post-hoc correlational estimates of a trained model's behavior rather than guarantees about its computation. Physics-informed deep learning offers a more direct route. Dabrowski et al.~\cite{dabrowski2023} developed a Bayesian physics-informed neural network that incorporated the level-set partial differential equation into the loss function to improve the model's performance on fire spread prediction. Similar methods incorporate the physical knowledge either as an additional input channel or as a soft constraint on the training objective. Both strategies leave the model free to ignore physics at inference time. Moreover, concatenated channels must compete with other features for influence. Additionally, a physics constraint written into a loss function is tied to one problem formulation and is less flexible when data modality changes. Conversely, a physical knowledge embedded in the architecture's computation can provide more flexibility as the scope of the problem and data changes. This motivates the research to embed fire behavior directly into the architecture with features such as directional relationship knowledge in wind and terrain. This is complemented with historically analogous fires retrieved by physics similarity, which both correct predictions and provide fire managers with past events that are physically comparable to the ongoing event. The second challenge is severe class imbalance. Benchmark datasets such as Next Day Wildfire Spread~\cite{huot2022} and TS-SatFire~\cite{zhao2025} document this challenge. Moreover, studies building on these benchmarks encounter the same imbalance issue~\cite{luo2026,han2026}. Both of the aforementioned datasets have active-fire pixels that typically constitute roughly 5\% of the labeled area. This can drive standard segmentation losses towards all-no-fire predictions. To address this, the research adapts a dual-stream gating component and applies it alongside the other two mechanisms on every backbone. The proposed mechanisms and their contributions are described below.

\begin{enumerate}
  \item \textbf{Wind/Slope Conditioned Attention Bias (WCAB/SCAB):}
  Wind and slope (which is calculated with elevation as seen in Section~\ref{sec:attention}) based directional priors have historically been key drivers to understand wildfire behavior ~\cite{rothermel1972}. Therefore, they are injected as additive terms into the pre-softmax attention computation through learnable zero-initialized scalars. The directional cue is thereby an identifiable component of the computation rather than an input channel or a soft loss penalty. It is a prescribed preference that the model can amplify, suppress, or reverse. The evaluation isolates the prior's contribution from the attention capacity that carries it through a control in which the same attention stage is present but the bias scalars are frozen at zero, so that the two are identical at initialization. While injecting priors as additive pairwise attention biases has been done before~\cite{chakraborty2026physics}, encoding directional fire-behavior cues in this way has not, to our knowledge, been examined for wildfire spread prediction.

  \item \textbf{Physics-Feature Retrieval-Augmented Output Correction (PFRAOC):}

  Analogous training tiles are retrieved from a memory bank keyed by a nine-dimensional physical description (Section ~\ref{sec:rag}) rather than by pixel similarity. The rationale is that a bank constructed on pixel similarity would be dominated by background pixels rather than active fires. Additionally, since the analogs are selected on physical similarity, the retrieved precedents are themselves physically comparable events. Historically analogous fire events are retrieved from a memory bank. This is computed by a physically meaningful state vector, as seen in Section~\ref{sec:rag}, rather than mere pixel similarity. The module then provides both a data-driven logit adjustment and historical precedents. This allows a fire behavior analyst to weigh the historical fires against the incident at hand. This approach is similar to case-based reasoning \cite{bannour2023emergency}; however, its distinguishing property is that the retrieval key's causal role is testable at inference. This is because corrupting a single physical component of the query, in an otherwise unchanged trained model, can reveal whether the correction depends on physically appropriate retrieval. 

  \item \textbf{Dual-Stream Gate:}
    A soft spatial gating mechanism that suppresses environmental
    feature channels in non-fire pixels, focusing the model's
    representational capacity on fire-adjacent regions where
    atmospheric and topographic conditions actively govern spread
    behavior.

\end{enumerate}

To assess the generality of the three proposed modules, they are evaluated across SwinUNETR~\cite{hatamizadeh2022swin}, MK-UNet~\cite{rahman2025mkunet},  Cross-Attentive Feature Interaction Module CNN (CAFIM-CNN) ~\cite{han2026}, U-Net~\cite{ronneberger2015,ayanzadeh2021improved,ayanzadeh2019cell,isensee2021,oktay2018attention,zhou2018unetpp,chen2021transunet}, and a Convolutional (Conv) Autoencoder (AE) ~\cite{lecun1989backpropagation}. 

The contribution of this study is a modular framework for examining how explicit domain cues interact with learned representations. For example, wind and slope enter attention as identifiable additive terms. Then, a separate correction module vectorized various fire features to retrieve similar next-day fire masks. The research investigates whether explicit directional priors and historical examples can make selected components of next-day wildfire prediction inspectable, while retaining predictive performance. This study has three objectives: (1) to separate the effects of fire-conditioned gating, attention capacity, and wind- and slope-conditioned biases;  (2) to assess the predictive contribution and input sensitivity of a physics-feature retrieval correction; and (3) to characterize how these effects vary across five host architectures and ensemble configurations. These objectives connect the need for examinable model components with the practical requirement that added complexity yield a defensible benefit. Performance is assessed using AUC-PR and F1, complemented by directional audits, retrieval perturbation, calibration measures, and computational costs. The evaluation criterion is a documented relationship between what each mechanism exposes, what it changes in the predictions, and the conditions under which its integration is beneficial. 

The remainder of the paper is organized as follows. Section~\ref{sec:related}
reviews related work in wildfire preparedness, segmentation methods across  domains, and physics-informed learning. Section ~\ref{sec:data} overviews the dataset from \cite{huot2022}.
Section~\ref{sec:methods} details the proposed gating, physical attention biases,
and retrieval mechanisms. Section~\ref{sec:results} presents experimental results,
followed by a discussion of limitations in Section~\ref{sec:discussion} and
conclusions in Section~\ref{sec:conclusion}.

% ============================================================
\section{Related Work}
\label{sec:related}
% ============================================================
Wildfire prediction supports decisions at several stages of an incident. Vision language models (VLMs) have been used for response \cite{esparza2025automated} and preparedness \cite{ayanzadeh2026wildfirevlm}. At the monitoring stage specifically, Ayanzadeh et al. (2026)  proposes WildfireVLM which combines satellite image fire and smoke detection with language-driven risk assessment to support early detection and response prioritization. Their study contextualizes risk assessment in the preparedness phase rather than mapping fire spread ~\cite{ayanzadeh2026wildfirevlm}. Other Pre-incident work includes ignition-probability mapping~\cite{pourmohamad2026predictive,tong2023mapping} and structure-level damage prediction with graph representations of fire transmission between vegetation and buildings~\cite{chulahwat2022integrated,chulahwat2024impact,esparza2026graph}. Closer to the present task, FireCast~\cite{radke2019} trained a compact CNN on weather rasters to forecast 24-hour burn perimeters around an active fire, Hodges and Lattimer~\cite{hodges2019} used deep convolutional networks to emulate FARSITE simulations at a small fraction of their cost, and R\"osch et al.\ (2024) trained a spatial graph neural network on 2016--2022 perimeter time series, reaching IoU values of 0.37 and 0.36 for Portugal and the wider Mediterranean region, respectively; these modest scores highlight the difficulty of next-day prediction even when the perimeter is known \cite{rosch2024data}. An advancement of next-day studies was made based on treating the problem as a segmentation task. Framing the problem as segmentation labels every pixel fire or no-fire at $t{+}1$, rather than collapsing the event to a single regional growth rate.

Segmentation methods have been used to solve complex problems in various domains, such as detecting medical diseases, monitoring structural cracking, and wildfire forecasting. Within the medical domain, segmentation architectures developed for biomedical imaging have been reused with little change for remote sensing and wildfire mapping. U-Net~\cite{ronneberger2015,ayanzadeh2021improved,ayanzadeh2019cell,isensee2021,oktay2018attention,zhou2018unetpp,chen2021transunet} suits dense geospatial prediction because its local receptive fields and multi-scale skip connections impose a spatial prior well matched to small, label-sparse tiles, and the self-configuring nnU-Net~\cite{isensee2021} showed that this recipe generalizes across tasks with minimal manual tuning. Attention gates on skip pathways~\cite{oktay2018attention}, nested skips~\cite{zhou2018unetpp}, and transformer encoders~\cite{chen2021transunet,valanarasu2021medt} extended the family, and Ghali et al. (2021)~\cite{ghali2021wildfire} applied TransUNet and MedT to forest-fire pixel segmentation. The same transfer is reflected here through SwinUNETR~\cite{hatamizadeh2022swin} and MK-UNet~\cite{rahman2025mkunet}. Within the material science domain, segmentation has been used to assess cracking and damaged concrete structures \cite{ccelik2025pixels}. For example, Zhang and Liu (2025) used segmentation network to enhance the identification of cracked edges by filtering out shadows, and applying a GAN network to the processed images \cite{zhang2025generative}. Lie et al. (2026) used various YOLO models to develop a pixel-level segmentation pipeline to detect various cracks, and they reached a segmentation accuracy rate of 68.4\%. They were able to increase the IoU value from 57.87 to 75.03 and surpassed state of the art models due to segmentation methods \cite{li2026pavement}. Wang et al. (2026) proposed an end-to-end framework that enhanced damage detection of reinforced concrete structures using segmentation methods, and achieved a mean IoU of 4.07\% \cite{wang2026framework}.

Within the wildfire domain, segmentation models and datasets continue to advance next-day spread prediction. Han et al.\ (2026) proposed FireSenseNet, a dual-branch CNN whose CAFIM fuses fuel and weather streams~\cite{han2026}. It reached an F1 score of 0.4176 and an AUC-PR of 0.3435 on Next Day Wildfire Spread. This is above the original benchmark values of 0.359 (F1) and 0.284 (AUC-PR) reported by Huot et al. (2022) ~\cite{huot2022}. Han et al. (2026) also showed how protocol choice distorts classification metrics on this task: re-evaluating CAFIM under the ``Both Days'' protocol of Luo et al. (2026) ~\cite{luo2026}, which uses the previous-day and next-day fire masks jointly as the target and relabels unobservable pixels as no-fire. This approach inflated its F1 score by 44.4\% with no change to the architecture, and by 44--50\% across the architectures they tested. Luo et al.\ (2026) reported an F1 score of 0.702 for a transform-domain fusion U-Net under that protocol, most of which is attributable to the previous-day mask being copied into the target~\cite{luo2026,han2026}. This finding informed the protocol adopted here (Section~\ref{sec:protocol}), which keeps the next-day target of Huot et al.~(2022) \ and reports AUC-PR alongside F1. Two further qualifications apply when comparing published scores. FireSenseNet's reported F1 was obtained with a decision threshold optimized on the test set~\cite{han2026}, whereas thresholds in this study are selected on training and validation data, so published values are not directly comparable with those reported here. Luo et al. (2026)\ also explored the TS-SatFire dataset of Zhao et al.~(2025) ~\cite{zhao2025}, a distinct multi-day dataset, obtaining an F1 score of 0.591 under the ``Both Days'' protocol. Zhao et al. (2025) original benchmark on the SwinUNET model was an F1 score of 0.374 ~\cite{zhao2025}. The dataset has robust spatial information for multiple day forecasting, but its training and test sets are considerably smaller than those of Huot et al. (2022); therefore, Next Day Wildfire Spread dataset is used in this research. Together, this body of work establishes a strong foundation for segmentation-based spread prediction while demonstrating that classification performance alone is an unstable basis for comparison across studies.

To enhance deep learning models, researchers have embedded physical knowledge into learning frameworks. The application of the knowledge varies. In physics-as-loss approach Vogiatzoglou et al.~(2025) constrains a PINN with mass and energy conservation to recover spread parameters such as advection velocity from observed fire evolution \cite{vogiatzoglou2025physics}. Wadhwani et al. (2025) explores a physics-as-data approach by training an LSTM surrogate on 64{,}000 high-fidelity grassland simulations, transferring physics through the training distribution itself \cite{wadhwani2025integrating}. From an architectural perspective, Abbas et al. (2026) implements FireCastFusion. This model appends a physics-guided diffusion layer to a temporal transformer for UAV-based fire-front forecasting. The physics acts as post-encoding refinement and a training regularizer, while the attention computation itself remains physics agnostic \cite{abbas2026firecast}. Outside the wildfire domain , PIBERT biases attention scores using PDE-residual diagnostics for CFD surrogates\cite{chakraborty2026physics}.This directs attention toward physically active regions, but without encoding the direction in which the physics drives the field. Additive attention biasing itself is a well-established design family, underpinning relative position encoding and representations \cite{wu2026flashbias}. However, mechanistically derived, directional biases remain unexplored for geospatial fire prediction. 

The methodological departure is to combine identifiable wind- and slope- based attention terms with a separately trained correction module that retrieves examples using a nine-dimensional environmental and fire-state descriptor. A fire conditioned gate provides an additional means for modulating environmental features according to the observed fire state. For convolutional hosts, an attention-capacity control is included so that the directional terms can be evaluated against an otherwise comparable attention-augmented model. The resulting design separates questions about added capacity, prescribed directionality, historical-example correction, and architectural compatibility. The contribution lies in this combined formulation and evaluation, building on established attention-bias and case-based prediction approaches while making the integration of their inspection capabilities explicit. The empirical test is whether these components expose useful inspection objects and improve, preserve, or reduce performance under clearly specified evaluation conditions.

% ============================================================
\section{Data}
\label{sec:data}
% ============================================================

\subsection{Dataset Description}

\begin{figure*}[t]
\centering
\includegraphics[width=0.92\textwidth]{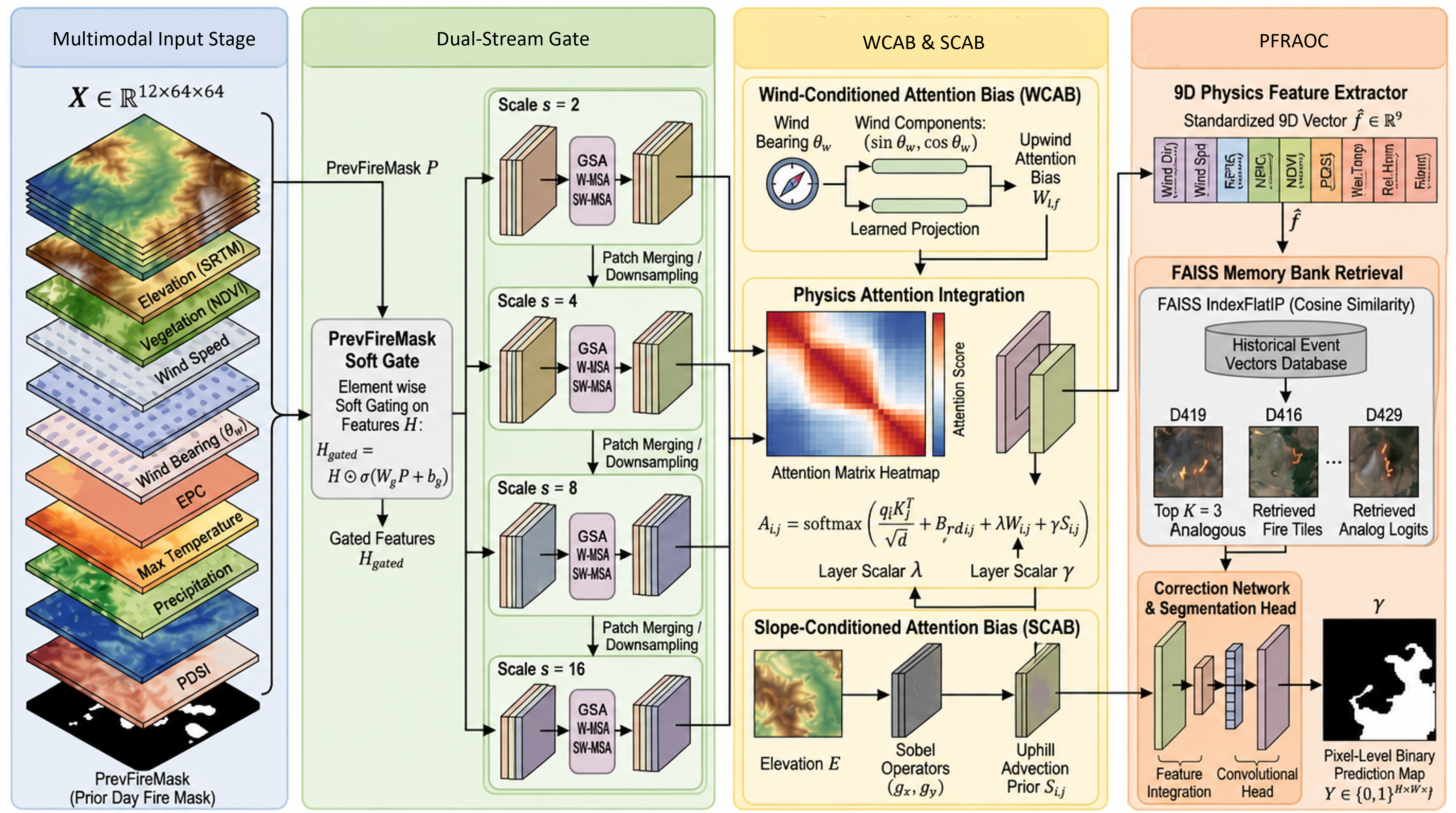}
\caption{The proposed mechanisms: (1) dual-stream gate; (2) WCAB, and SCAB; (3) PFRAOC. These can be combined together as depicted (the order of the steps is arbitrary) or in modular format as Section \ref{sec:results} show. The first module shows how the gating is used based on previous fire mask (PrevFireMask). The second module demonstrates how the attention calculation is augmented with wind and slope attention biases. The third module shows how a 9D vector is used to create a memory bank for output corrections during inference.}
\label{fig:workflow}
\end{figure*}

This study uses the Next Day Wildfire Spread
dataset, originally developed by Huot et al.~(2022) ~\cite{huot2022}. This
dataset was collected using Google Earth Engine and includes 18,545
spatial samples at 1 km resolution, comprising 14{,}979 train, 1{,}877
val and 1{,}689 test tiles. This multivariate repository aggregates
nearly a decade of remote sensing data from 2012 to 2020 across the
contiguous United States. Each spatial sample represents a 64 km by 64 km
geographical region mapped at 1 km resolution during an active fire
event. This dataset overlays daily binary fire masks indicating ``fire''
or ``no fire'' alongside a robust set of explanatory variables. These
comprehensive features include crucial environmental drivers such as
topography~\cite{farr2007}, weather conditions~\cite{abatzoglou2013},
drought indices~\cite{abatzoglou2014}, vegetation
health~\cite{didan2018}, and human population
density~\cite{ciesin2018}. This
combination of 2D spatial fire mapping and multi-dimensional
environmental constraints provides a highly feature-rich foundation for
training machine learning models to forecast fire-spread behavior with a 24-hour lead time. Fig.~\ref{fig:workflow} demonstrates how these channels enter the model.

Each sample stacks twelve co-registered input channels
on the common \SI{1}{\kilo\meter} grid: topography (elevation),
vegetation (NDVI), six meteorological fields (maximum and minimum
temperature, wind speed, wind direction, specific humidity,
precipitation), drought (PDSI), fuel dryness (energy release component,
ERC), human exposure (population density), and the previous-day fire
mask. Their source products range from \SI{30}{\meter} terrain data to
approximately \SI{4}{\kilo\meter} meteorological fields, with every
channel resampled to the common grid. The label is the next-day fire mask from the MODIS MOD14A1
daily \SI{1}{\kilo\meter} thermal-anomaly composite~\cite{giglio2015},
which defines the grid every other channel is resampled to, and encodes
fire, no fire, and an unobservable class ($-1$) for cloud-occluded or
unprocessed pixels that is withheld from both the training loss and
every reported metric. The fixed split assigns whole weeks of the
2012--2020 record to the three partitions in an $8{:}1{:}1$ ratio, with
a one-day buffer between weeks.

\subsection{Dataset Class Distribution}

Class imbalance is the benchmark's defining difficulty.  A pixel-frequency audit of the
released next-day masks reports \SI{0.98}{\percent} fire,
\SI{97.26}{\percent} no fire, and \SI{1.77}{\percent}
unobservable~\cite{taylor2024}: a negative-to-positive ratio near
$99{:}1$ once $-1$ pixels are dropped, which also thins effective label
density by a further \SI{1.8}{\percent}. Independent audits put next-day
spread under \SI{1}{\percent}~\cite{han2026}.
Sparsity is also uneven across tiles. The fire grows only in \SI{58}{\percent} of samples (10{,}798), contracts in
\SI{39}{\percent} (7{,}191), and is unchanged in the
remainder~\cite{huot2022}. This leaves nearly half the corpus with little or
no net new-spread signal. Accuracy is consequently uninformative, since
a degenerate all-no-fire predictor exceeds \SI{99}{\percent}, and
AUC-ROC is dominated by the true-negative pool. AUC-PR is the primary
metric here, as in the original benchmark.

% ============================================================
\section{Methodology}
\label{sec:methods}
% ============================================================

This research proposes three architectural components for next-day wildfire spread prediction that selected domain-informed features for inspection. To determine whether these components are transferable across different models, a suite of architectures is tested, as described in Section~\ref{sec:model_arch}. Once the architectures are individually augmented with all three mechanisms, the research explores how an ensemble of architectures interact with the mechanisms. The first mechanism is attention augmentation.~A pair of physics-conditioned attention biases, (1) Wind-Conditioned Attention Bias (WCAB) and (2) Slope-Conditioned Attention Bias (SCAB), are injected directly into the spatial attention engine, described in Section~\ref{sec:attention}. The second is a physics-feature retrieval-augmented output correction (PFRAOC) that corrects predictions using retrieved training tiles, explored at both training and inference,  as seen in Section~\ref{sec:rag}. The third is a dual-stream gate that concentrates the model's representational capacity on fire-adjacent regions, described in Section~\ref{sec:gating}. ~Together, these components make selected directional assumptions and retrieved precedents identifiable within the prediction pipeline. Section \ref{sec:results} explores whether these mechanisms enhance the predictive performance of the five architectures.  As mentioned, the three mechanisms are explored when multiple architectures are used for predictions. The evaluation protocol for individual and ensembled models, the target definition, and the use of each data split are summarized in Section \ref{sec:protocol}. 

\subsection{Model Architectures}
\label{sec:model_arch}

\begin{figure*}[t]
\centering
\includegraphics[width=\textwidth]{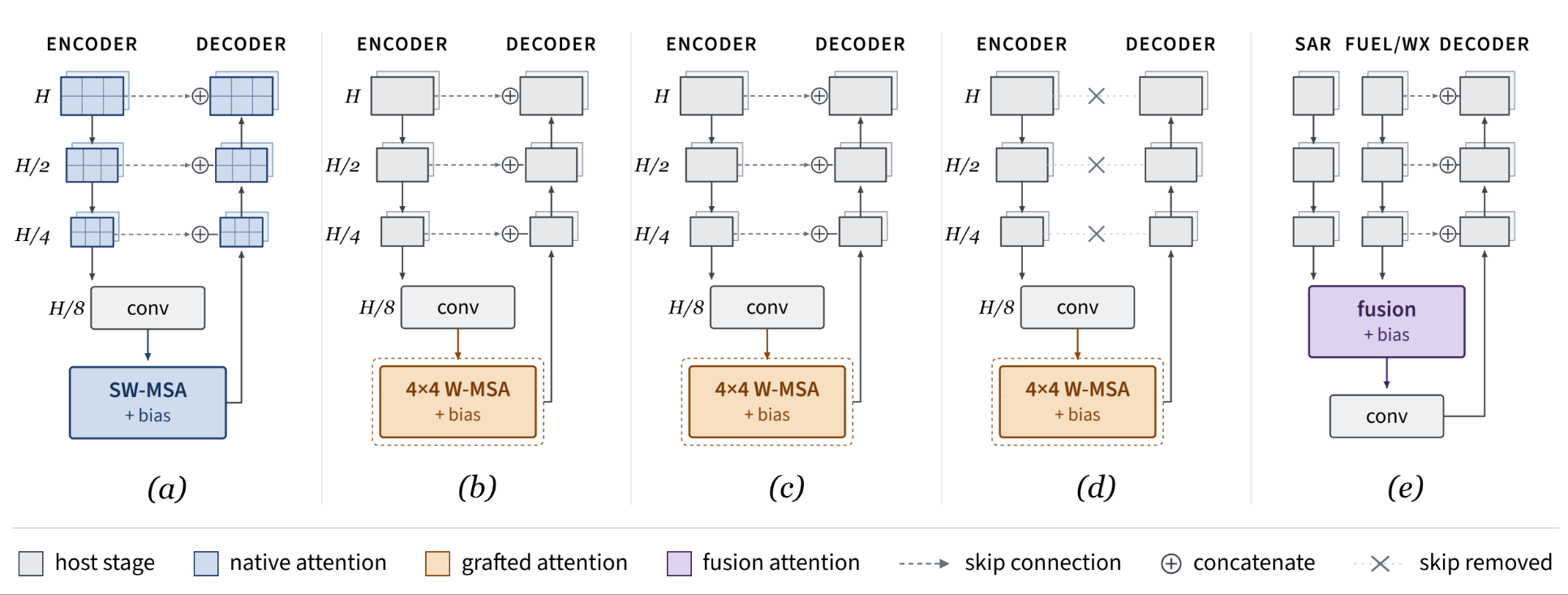}
\caption{(a) Swin UNETR, (b) MK-UNet, (c) plain U-Net, (d) Conv-AE, and (e) CAFIM-CNN. They are drawn to the three controlled contrasts (context range, skip connections, joint vs.\ split inputs) and to the physics-attention entry point. Specs and parameter counts is in the Table~\ref{tab:host_panel}.}
\label{fig:host_panel}
\end{figure*}

To test whether the three proposed components can be integrated across architectures, they are attached to five hosts that independently vary context range, skip connections, and modality fusion (Fig.~\ref{fig:host_panel}, Table~\ref{tab:host_panel}), then scored on the same ablation ladder (Table~\ref{tab:ablation}). SwinUNETR is the only host with native windowed attention. The four convolutional hosts receive a grafted $4\times 4$ windowed-attention stage in the encoder so that WCAB/SCAB can be applied identically. CAFIM-CNN is the only host with early fuel--weather separation, so it tests whether a directional attention prior is redundant with an existing cross-modal fusion block.

\begin{table*}[t]
\centering
\caption{Specification of the five host architectures in Fig.~\ref{fig:host_panel}.}
\label{tab:host_panel}
\small
\setlength{\tabcolsep}{3pt}
\renewcommand{\arraystretch}{1.25}
\newcommand{\nativepath}{\begingroup\setlength{\fboxsep}{2.2pt}%
  \colorbox{paperblue!12}{\textcolor{paperblue}{\textbf{\scriptsize Native}}}\endgroup}
\newcommand{\graftedpath}{\begingroup\setlength{\fboxsep}{2.2pt}%
  \colorbox{paperorange!12}{\textcolor{paperorange}{\textbf{\scriptsize Grafted}}}\endgroup}
\newcommand{\fusionpath}{\begingroup\setlength{\fboxsep}{2.2pt}%
  \colorbox{paperviolet!12}{\textcolor{paperviolet}{\textbf{\scriptsize Fusion}}}\endgroup}
\newcommand{\yskip}{\textcolor{papergreen}{\checkmark}}
\newcommand{\nskip}{\textcolor{paperred}{--}}
\begin{tabularx}{\textwidth}{@{}
  >{\raggedright\arraybackslash}p{2.80cm}
  >{\centering\arraybackslash}p{1.35cm}
  >{\raggedright\arraybackslash}p{2.50cm}
  >{\centering\arraybackslash}p{0.95cm}
  >{\centering\arraybackslash}p{1.05cm}
  >{\centering\arraybackslash}p{1.40cm}
  >{\raggedright\arraybackslash}X@{}}
\toprule
\textbf{Host} &
\textbf{A / D} &
\textbf{Context} &
\textbf{Skips} &
\textbf{Inputs} &
\textbf{Path} &
\textbf{Purpose} \\
\midrule
\textbf{SwinUNETR}~\cite{hatamizadeh2022swin}
  & 6.30 / 6.30
  & Shifted-window, long-range
  & \yskip
  & Joint
  & \nativepath
  & Physics prior on an existing attention graph \\
\addlinespace
\textbf{MK-UNet}~\cite{rahman2025mkunet}
  & 0.32 / 0.40
  & Multi-kernel, local
  & \yskip
  & Joint
  & \graftedpath
  & Transfer to an ultra-light convolutional host \\
\addlinespace
\textbf{U-Net}~\cite{ronneberger2015}
  & 1.93 / 2.46
  & Hierarchical conv., local
  & \yskip
  & Joint
  & \graftedpath
  & Conventional skip-connected control \\
\addlinespace
\textbf{Conv-AE}
  & 1.74 / 2.27
  & Local conv.; no recycling
  & \nskip
  & Joint
  & \graftedpath
  & Skip-free, minimal spatial prior \\
\addlinespace
\textbf{CAFIM-CNN}~\cite{han2026}
  & 3.02 / 4.34
  & Cross-modal fusion, local
  & \yskip
  & Split
  & \graftedpath
  & Collision with an existing fuel--weather prior \\
\bottomrule
\end{tabularx}
\vspace{0.22em}
{\footnotesize\textit{Note.}
Each row is one backbone used in every later experiment. SwinUNETR already has windowed attention; the four convolutional hosts receive a grafted $4\times 4$ attention stage so that WCAB/SCAB can be applied in the same way. Parameter counts are the unmodified backbone (A) and the full gate plus WCAB/SCAB stack (D).}

\end{table*}

\subsection{Physics-Conditioned Attention Bias}
\label{sec:attention}
Standard spatial self-attention is agnostic to physical advection and topographic gradients. Attention weights are learned purely from data, with no structural preference for the direction in which fire is physically driven to spread. To address this, the research augments the spatial attention scores with physics-derived
directional biases for wind and slope. The rationale is to introduce directional preferences motivated by the roles of wind and slope in fire behavior~\cite{rothermel1972} as identifiable terms in the attention computation. The terms do not implement a Rothermel rate-of-spread model, and they do not force the complete attention distribution or prediction to follow the prescribed direction. Rather, their learned signed coefficients can suppress or reverse the prior. This is what makes the sign and magnitude of the learned coefficients an empirical finding rather than a property guaranteed by construction.

Because these biases are added directly to a self-attention score matrix, applying them to an architectural backbone involves two additions. The first is the self-attention mechanism itself. SwinUNetR has this attention mechanism built in already. Conversely, the remaining models are convolutional hosts, so a small windowed-attention module has to be added to each one instead. It processes the feature map in $4\times4$ patches, alternating between fixed grid of windows and a shifted one so information can pass between neighboring patches. This module is inserted partway through the encoder rather than at the input or output. Section~\ref{sec:WCAB} and  Section ~\ref{sec:SCAB} describes wind ($W_{ij}$) and -slope ($S_{ij}$) bias term, respectively. These terms are added to that stage's pre-softmax attention score. This creates a potential confound, a performance gain from adding WCAB/SCAB could reflect the physics prior itself or the added representational capacity of the attention stage that it carries. Therefore, the ablation design includes a control variant in which the identical attention stage is present, but the bias scalars are frozen at zero. This way, the physics contribution is measured net of the attention graph.

\subsubsection{Wind-Conditioned Attention Bias (WCAB)}
\label{sec:WCAB}
Wind direction is encoded in the dataset as a normalized meteorological
bearing $\theta_w$. This bearing is decomposed into orthogonal Cartesian
components ($\sin\theta_w$, $\cos\theta_w$) to avoid circular
discontinuities. Because the host architectures compute attention at
different spatial scales (downsampling factors $s \in \{2, 4, 8, 16\}$,
with the schedule depending on where the attention stages sit in each
backbone as seen in Section~\ref{sec:model_arch}), the component maps are
average-pooled to each scale and cyclically shifted to align with the
shifted attention window partitions. Within each window, the per-window
mean components $(\overline{\sin\theta}_w, \overline{\cos\theta}_w)$
are computed.

The wind attention bias $W_{ij}$ between tokens $i$ and $j$ is then
defined as:
\begin{equation}
  W_{ij} = \Delta x_{ij} \overline{\sin\theta}_w - \Delta y_{ij} \overline{\cos\theta}_w,
  \label{eq:wcab}
\end{equation}
where $\Delta x_{ij} = \operatorname{col}_j - \operatorname{col}_i$ and
$\Delta y_{ij} = \operatorname{row}_j - \operatorname{row}_i$ represent
the horizontal (east--west) and vertical (north--south) coordinate
displacements between the tokens within the window. Under the dataset's
image-coordinate convention, $W_{ij} > 0$ when token $j$ lies downwind
of token $i$ --- along the axis of expected fire advection. Adding this
bias to the attention score directs each token's attention toward the
direction in which fire is physically driven to spread. The bias enters the attention score through a
learnable per-layer scalar $\lambda$, initialized to zero.

\subsubsection{Slope-Conditioned Attention Bias (SCAB)}
\label{sec:SCAB}

Terrain slope is a fundamental driver of wildfire spread rate.
As fire burns uphill, the flame tilts toward the unburned vegetation
above it, preheating the fuel, shortening the ignition distance, and
accelerating propagation; this effect is quantified in the Rothermel
model through the slope correction factor
$\varphi_{\mathrm{slope}}$~\cite{rothermel1972}.
Conversely, fire burning downhill must project heat against gravity
and away from unburned fuel, slowing its rate of spread.
Standard self-attention is spatially symmetric. A token at the base
of a slope attends to tokens uphill and downhill with equal weight,
with no awareness of which direction fire is likely to advance. SCAB introduces the uphill--downhill asymmetry of terrain-driven spread as a directional preference in the attention score. 

To compute the terrain gradient, the elevation channel $\mathbf{E}$
is processed using a separable $3\times3$ Sobel operator:
\begin{equation}
  g_x = \tfrac{1}{8}\,\mathrm{Sobel}_x(\mathbf{E}),
  \qquad
  g_y = \tfrac{1}{8}\,\mathrm{Sobel}_y(\mathbf{E}),
  \label{eq:sobel}
\end{equation}
where $g_x$ and $g_y$ are the east--west and north--south components
of the terrain gradient, respectively.
The $\tfrac{1}{8}$ factor normalizes the Sobel kernel so that $g_x$ and $g_y$ are expressed in elevation units per pixel; because the grid spacing is fixed at \SI{1}{\kilo\meter} in this dataset, the gradient is proportional to the physical slope, but the factor by itself does not make the bias independent of spatial resolution.

At each attention layer, the gradient maps are downsampled to the
current feature resolution and averaged within each attention window
to yield per-window mean gradients $(\bar{g}_x, \bar{g}_y)$.
The slope attention bias between token $i$ and token $j$ is then:
\begin{equation}
  S_{ij} = \Delta x_{ij}\,\bar{g}_x + \Delta y_{ij}\,\bar{g}_y,
  \label{eq:scab}
\end{equation}
where $\Delta x_{ij} = \mathrm{col}_j - \mathrm{col}_i$ and
$\Delta y_{ij} = \mathrm{row}_j - \mathrm{row}_i$ are the spatial
displacements between tokens.
Geometrically, $S_{ij}$ is the dot product of the displacement vector
from $i$ to $j$ with the local terrain gradient. It is positive when $j$ lies uphill of $i$ zero, when the two tokens are at the same elevation, and negative when $j$ lies downhill. The bias therefore parameterizes the uphill axis—the direction toward which terrain-driven fire preferentially advances. Its contribution to the attention score is controlled by a learnable signed scalar $\gamma$ per attention layer, initialized to zero, which allows the model to amplify, suppress, or invert the slope prior depending on whether terrain is the dominant spread driver at that layer. For example, it may be suppressed in flat or wind-dominated fire environments. The magnitude and sign of the converged $\gamma$ values are reported in Section~\ref{sec:interpretability}.

\subsubsection{Attention Score Integration}
The physical biases are integrated into windowed self-attention as
additive terms on the pre-softmax attention scores. In the SwinUNETR
host this is the native Swin window attention. Conversely, in convolutional hosts,
which lack an attention-score matrix, the identical formulation is
carried by lightweight windowed self-attention blocks inserted at
intermediate encoder stages (Section~\ref{sec:model_arch}). In either
case the attention matrix is computed as:
\begin{equation}
  \begin{split}
  A_{ij}
    &= \operatorname{softmax}\!\left(
       \frac{q_i k_j^\top}{\sqrt{d}} + B_{\mathrm{rel},ij}\right. \\
    &\qquad\left.{}+ \lambda W_{ij} + \gamma S_{ij}
       \right),
  \end{split}
  \label{eq:full_attn}
\end{equation}
where $q_i, k_j$ are the query and key vectors, $d$ is their dimension,
$B_{\mathrm{rel},ij}$ is the standard relative position bias, and
$\lambda, \gamma \in \mathbb{R}$ are the learnable per-layer scaling
coefficients introduced above. Crucially, both $\lambda$ and $\gamma$
are initialized to zero: at initialization each host is functionally
identical to its physics-free counterpart, and the biases
activate only if they reduce the training loss. This zero-initialization
property also underpins the ablation design in
Section~\ref{sec:results}, where a control variant with $\lambda$ and
$\gamma$ frozen at zero isolates the contribution of the directional
priors from that of the added attention capacity itself.

A useful property of Eqs.~(\ref{eq:wcab}) and~(\ref{eq:scab}) follows from their form. Writing the per-window mean directional vector as $\mathbf{a}$ and the token positions as $\mathbf{r}(\cdot)$, each bias is $b_{ij} = \mathbf{a}\cdot[\mathbf{r}(j) - \mathbf{r}(i)]$. Under the row-wise softmax over keys $j$, the query-dependent term $-\mathbf{a}\cdot\mathbf{r}(i)$ is constant across keys and cancels, so the additive term acts as a window-level preference over key positions in the direction of $\mathbf{a}$. The complete attention remains query dependent through the query--key scores and the relative-position bias. The bias is therefore best described as an inspectable directional preference imposed on the attention scores, not as a mechanism that enforces advection. Because the wind bias is built from the same wind input against which it is later audited (Section~\ref{sec:interpretability}), a high alignment between the two is largely a property of the construction; the audit verifies that the learned scalars preserved the prescribed orientation rather than demonstrating that the model discovered wind-driven spread.

\subsection{Physics Feature Retrieval-Augmented Correction}
\label{sec:rag}
Augmenting a model's predictions with retrieved reference cases has precedent in case-based reasoning, which has supported emergency management decision-making~\cite{bannour2023emergency}, and in analog forecasting, which uses historically similar atmospheric states as a predictive prior~\cite{lorenz1969atmospheric,vandendool1994searching}. Additionally, Retrieval-Augmented Generation (RAG) has proven effective in
knowledge-intensive tasks by augmenting model predictions with relevant
information retrieved from a memory bank~\cite{lewis2020}. The mechanism proposed here is an instantiation of this lineage rather than a departure from it. Standard learned retrieval 
approaches encode inputs as dense learned embeddings and perform
similarity search in that embedding space. Applied directly to wildfire
tiles, however, such embeddings would be dominated by the abundant
non-fire background, and the resulting similarity space would be
opaque, as a practitioner could not inspect why two fires were
deemed analogous.

To address this, the research replaces learned embeddings with a
nine-dimensional physics feature vector computed from domain-relevant physical quantities. This vector is computed once per tile as a preprocessing step and serves as the basis for the memory bank. Using interpretable features from the dataset rather than opaque embeddings makes the bias on which tiles are retrieved inspectable. However, it does not by itself guarantee that a retrieved tile is an appropriate analog for a given fire. 

Seven scalar features are derived from the $64 \times 64$ geospatial
input channels by computing the spatial mean across all pixels: (1)
wind speed, (2) Energy Release Component (ERC), (3) elevation, (4)
Normalized Difference Vegetation Index (NDVI), (5) Palmer Drought
Severity Index (PDSI), (6) maximum temperature, and (7) the fraction of currently burning pixels. This is simply the share of grid cells whose previous-day fire mask exceeds 0.5, so the retrieval key knows how large the active fire already is, not only the weather and terrain around it. The remaining two features are the Cartesian
wind-direction components of Section~\ref{sec:attention}.

This forms the full nine-dimensional physics feature vector
$\mathbf{f}$:
\begin{equation}
  \begin{aligned}
  \mathbf{f} = \bigl[&\cos\theta,\; \sin\theta,\; \bar{v}_w,\;
    \overline{\mathrm{ERC}},\; \bar{z}, \\
    &\overline{\mathrm{NDVI}},\; \overline{\mathrm{PDSI}},\;
    \bar{T}_{\max},\; f_{\mathrm{fire}}\bigr]^\top
    \in \mathbb{R}^9,
  \end{aligned}
  \label{eq:physics_vector}
\end{equation}
where $\bar{(\cdot)}$ denotes the spatial mean over the $64\times64$
tile and $f_{\mathrm{fire}}$ is the fraction of burning pixels.

To ensure each physical property contributes equally to the similarity
computation regardless of its original scale or units, each feature
dimension is standardized using global statistics computed over the
full training set:
\begin{equation}
  \tilde{f}_k = \frac{f_k - \mu_k}{\sigma_k},
  \quad k = 1, \ldots, 9,
  \label{eq:zscore}
\end{equation}
where $\mu_k$ and $\sigma_k$ are the training-set mean and standard
deviation of the $k$-th feature dimension. The standardized vector is
subsequently L2-normalized to unit length:
\begin{equation}
  \hat{\mathbf{f}} = \frac{\tilde{\mathbf{f}}}{\|\tilde{\mathbf{f}}\|_2}.
  \label{eq:l2norm}
\end{equation}
The normalized physics vectors are indexed with Facebook AI Similarity
Search (FAISS)~\cite{johnson2019billion} using the \texttt{IndexFlatIP}
(flat inner-product) index, which performs exact search without index
training or compression and supports GPU acceleration to mitigate
retrieval latency. Since each vector is L2-normalized
(Eq.~\ref{eq:l2norm}), the inner product between any two vectors equals
their cosine similarity, mapping the similarity ranking onto $[-1, 1]$.
The memory bank spans all training and validation tiles
($14{,}979 + 1{,}877 = 16{,}856$ entries), providing access to the full
labeled distribution at retrieval time. When queries are drawn from the
bank itself, as during training of the correction module, any
neighbor whose index matches the query tile is excluded to prevent
self-retrieval; test queries are never bank members. Unless otherwise
stated, $K = 3$ analogs are retrieved throughout. Because the bank is drawn from the training and validation distribution, retrieved tiles may come from incidents related to the query, and excluding the query's own row does not establish independence from related tiles. Section \ref{sec:rag_results} reports a wrong-physics ablation in which the query's wind component are negated at inference so that tiles with opposite wind conditions are retrieved. This ablation tests sensitivity to the retrieval descriptor. It does not rule out dependence on related incidents or establish generalization to a chronologically independent bank, a limitation future research can mitigate. Additionally, the bank's next-day fire masks are sparse (median fire fraction 0.4\%), which limits what the retrieved content itself physically contributes. The supplementary analysis finds that the correction's gain is statistically unchanged under random or fixed retrieval. This indicates the gain seen in the results section is attributable to the added correction capacity rather than to the specific analogs selected. Nevertheless, the retrieved analogs remain measurably closer physical matches to the observed spread than random tiles, supporting their role as inspection objects. Therefore, there is no accuracy trade off, but interpretability remains physically accurate. Curating a denser bank so that retrieval content can also contribute to accuracy is a remaining step to mitigate this second limitation.

The retrieved analogs are used in two configurations. In the primary,
post-hoc configuration, the $K$ nearest analogs to the query tile's
physics descriptor are retrieved at inference; their observed next-day fire masks are averaged into a single $64 \times 64$ prior map.  No spatial registration between the query and the retrieved tiles is implied by this average, so the map acts as a retrieval-derived tile-coordinate prior that a lightweight convolutional correction network transforms. The result is 
scaled by a learnable weight $\gamma_r$, and added to the logits of the
frozen base model. This provides incident commanders with a
calibrated prediction alongside the retrieved tiles used to compute it, which remains available for inspection. In the second, in-loop configuration, the analog
prior instead enters training as a consistency loss with weight
$\lambda_r$ (using $K =3$), so that the deployed model is a single
forward pass with no retrieval index. The two configurations are
compared in Section~\ref{sec:rag_results}. Relative to its case-based and analog forecasting lineage, the mechanism's distinguished properties are threefold: (1) the correction's sensitivity to the retrieval descriptor is testable at inference through the wrong-physics ablation, (2) the correction grafts onto a frozen deep segmentation model without retraining, and its attachment is demonstrated on five architecturally distinct hosts, and (3) the retrieved tiles serve simultaneously as the correction signal and as inspection objects surfaced to the practitioner. 

\subsection{Gating Mechanism}
\label{sec:gating}

The proposed gate is a dual-stream design applied at the input stage, before the backbone: the Previous Fire Mask channel is projected into a dedicated fire-state feature stream that passes through ungated, while the remaining environmental channels are modulated by a spatial weight map
conditioned on that fire state. This preserves gradient flow throughout the spatial grid while concentrating representational weight near the
active fire. The gating operation is:
\begin{equation}
  \mathbf{H}_{\mathrm{gated}} = \mathbf{H}
    \odot \sigma(\mathbf{W}_g \mathbf{P} + b_g),
  \label{eq:gate}
\end{equation}
where $\mathbf{H}$ is the projected environmental feature
representation, $\sigma$ is the sigmoid activation function,
$\mathbf{W}_g$ and $b_g$ are learnable parameters, and
$\mathbf{P}$ is the Previous Fire Mask channel encoding the previous day's fire active fire masks. The sigmoid produces a spatial weight map in
$(0, 1)^{H \times W}$. Weights can approach but never equal zero, so even strongly attenuated pixels retain a weak gradient signal, allowing the model to learn how field conditions relate to eventual spread. Pixels within and adjacent to the active fire can receive weights approaching unity, preserving their environmental features at full magnitude. Conversely, distant unburned pixels are attenuated. Because $\mathbf{W}_g$ and $b_g$ are learned, the spatial pattern and strength of this attenuation are determined during training rather than fixed by a hand-tuned threshold. Because the gate operates on input projections and requires nothing of
the backbone's internals, it can be attached to each of the five hosts without modifying the backbone. The degree to which each host benefits from it and the extent to which the learned gate exploits its spatial freedom is examined empirically in Section~\ref{sec:results}.

\subsection{Evaluation Protocol}
\label{sec:protocol}

The primary target is the observed active-fire mask on day $t{+}1$, including pixels that were also active on day $t$. Pixels with unknown day-$t{+}1$ labels are excluded from the loss and from every reported metric. This target differs from both newly detected fire outside the prior-day active-fire mask and the union of the two daily masks. The same target and valid-pixel mask are applied to every model, table, and figure.
All models are evaluated on the Huot et al. (2022) ~\cite{huot2022} test set under the same evaluation protocol applied to the examined architectures, mechanism configurations, and ensemble. The decision threshold is selected as the F1-optimal point on the combined training and validation sets. The threshold is then frozen before the model is applied to the held-out test set, so it is never tuned on the data it is evaluated on. AUC-PR is the primary metric because it is threshold independent. Uncertainty is quantified in two ways that capture different sources of variation and are not interchangeable. First, every point estimate carries a 95\% confidence interval from $N{=}1{,}000$ tile-level bootstrap resamples of the test set, reflecting sampling variability within one trained model. Second, because bootstrap resampling cannot capture variability introduced by training itself, the ablation results are retrained and re-evaluated across three seeds, and the mean AUC-PR and F1 across the three seeds are reported with their standard deviation. The bootstrap intervals are computed for a single seed and therefore describe resampling uncertainty of one trained model; they are not intervals for the three-seed mean, and overlapping intervals between configurations do not establish equivalence or the absence of an effect. Paired resampling of per-tile differences between configurations, which would assess incremental gains directly, is not reported here, so small differences between configurations should be read as estimates with uncertainty. This protocol is the same for individual and ensembled models. Backbone weights are fit on the training split; descriptor standardization statistics (Eq.~\ref{eq:zscore}) are computed on the training split; the retrieval bank and the fitting of the correction network draw on training and validation tiles with self-retrieval excluded; the decision threshold is selected on the combined training and validation sets; and the test split is used only for final evaluation.

Beyond classification, the models are evaluated from a cost and calibration perspective. Cost is reported as parameter count, single-tile GPU latency, and peak memory at inference. These are independent deployment constraints that capture a model's storage and distribution footprint, its inference throughput, and the hardware required to run it. For calibration, it is reported as Expected Calibration Error (ECE) and Brier score. ECE bins predictions by confidence and measures the weighted average gap between predicted probability and observed outcome frequency within each bin. Under this evaluation it summarizes the average gap between stated probability and observed frequency. It does not by itself establish that a probability can be used at face value in a new incident. Brier score is the mean squared error between predicted probability and outcome. It is a proper scoring rule that penalizes predictions that are calibrated but uninformative. Lower values are better for both metrics. This is because the Brier score measure overall probabilistic quality rather than calibration alone, the two are reported together. Cost and calibration let any gain in classification be weighted against the computational resources and trustworthiness of the probabilities the model outputs.

%==========================================================================
\section{Experiments and Results}
\label{sec:results}
The study objectives are: (1) to separate the effects of fire-conditioned gating, added attention capacity, and wind/slope conditioned biases; (2) to assess the predictive contribution and input sensitivity to physics-feature retrieval correction; (3) to characterize how these effects vary across five host architectures and their ensemble configurations. Together, these objectives connect the need for examinable model components with the practical requirement that added complexity yield a defensible benefit, and the results address each in turn. Section \ref{sec:gate_attn} isolates the gating and attention augmentation components, first individually and then in conjunction across the five models (objective ~1). Section \ref{sec:interpretability} examines how the physical inputs shape the learned attention biases and the extent to which wind , slope, and fire spread can be audited from them. Section \ref{sec:rag_results} evaluates the PFRAOC correction, including its physical contribution and its sensitivity to the retrieval inputs (objective ~2). Section \ref{sec:ens_results} characterizes how each mechanism's effect varies across the individual architecture and their ensemble configuration (objective ~3). These sections quantify gains with AUC-PR and F1. Section ~\ref{sec:cal_cost_results} complements them with calibration and computational-cost results for all model and ensemble configurations.

\subsection{Attention Augmentation and Dual Stream Gate Results}
\label{sec:gate_attn}

To assess whether the proposed methods transfer across backbones rather than being artifacts of a single one, Table~\ref{tab:ablation} reports the full ablation of the gate and the directional bias mechanisms (WCAB and SCAB) on five models. These are: (1) SwinUNETR, (2) MK-UNet, (3) CAFIM-CNN, (4) U-Net, and (5) Conv-AE. Each cell reports the three-seed mean with, in brackets, 95\% confidence interval from $N=1{,}000$ tile-level bootstrap resamples of a single seed (Section \ref{sec:protocol}). Column A reports each backbone without any of the mechanisms. The Conv-AE has the lowest F1 and AUC-PR score (F1 = 0.396, AUC-PR = 0.327), possibly due to the lack of skip connections that recover fine spatial details. SwinUNETR has similar metrics (F1 = 0.407, AUC-PR = 0.334). This could be due to the transformer backbone not having a convolutional locality prior nor built-in modality separation. The remaining three models are rather similar to each other as the U-Net (F1 = 0.415, AUC-PR = 0.346) and MK-UNet (F1 = 0.411, AUC-PR = 0.345) benefit from convolutional locality with skip connections. The CAFIM-CNN has the highest classification metrics (F1 = 0.416, AUC-PR = 0.356) in column A due to the early separation of fuel and weather modalities. When the gating component is introduced (column B), the benefit is backbone-conditional. SwinUNETR improves substantially (F1: 0.418; AUC-PR: 0.358). Conversely, the U-Net, Conv-AE, and MK-UNet improve modestly as their AUC-PR only increased by 0.006, 0.008, 0.002, respectively. The CAFIM-CNN showed a small degradation as its AUC-PR went from 0.356 to 0.355. These changes are small and lie within the bootstrap intervals. A plausible interpretation, which the present experiments do not isolate, is that the gate helps backbones that lack an intrinsic means of prioritizing fire-adjacent signal and is redundant for hosts whose architecture already impose such structure, such as MK-UNet (spatial locality) and CAFIM-CNN (fuel and weather separation).

The attention augmentation described in Section~\ref{sec:attention} bundles two mechanisms that could each enhance prediction: (1) the self-attention operation itself and (2) the wind- and slope-derived bias terms. To isolate where any benefit stems from, column C reports a control in which the identical attention stage is present but the wind and slope scalars are frozen at zero. Additionally, column C has the gating mechanism. For SwinUNETR, where attention is native to the backbone, this configuration is architecturally identical to column B, and the measured scores confirm the equivalence (F1 = 0.417, AUC-PR = 0.359). On all four convolutional hosts, the grafted attention stage consistently reduces performance relative to column B ($\Delta$AUC-PR: MK-UNet $-0.009$, CAFIM-CNN $-0.005$,  U-Net $-0.008$, Conv-AE $-0.010$), demonstrating that added generic attention capacity is not itself the source of any subsequent gain.

Column D reports the full configuration, adding the directional biases to the column C attention stage and the gating configuration. Since column C and D are identical at initialization, the C$\to$D contrast attributes any difference specifically to the  directional form of the bias. The prior delivers its clearest gains on SwinUNETR, which attains the best overall scores (F1 = 0.421, AUC-PR = 0.364). MK-UNet not only recovers but exceeds the graft cost ($\Delta$AUC-PR = $+0.014$ over column C). The U-Net and Conv-AE show no reliable change over their controls. The CAFIM-CNN model declines slightly ($\Delta$AUC-PR = $-0.004$). One hypothesis, not isolated here, is that its built-in-cross modal fusion already encodes the fuel--weather interactions the prior supplies. Taken together, the pattern indicates that the directional bias is not a universal additive. It is utilized best when the backbone either computes attention natively (SwinUNETR) or integrates the grafted stage efficiently at very small scale (MK-UNet). In the other architectures, the attention augmentation slightly reduces classification performance while adding a directional prior whose orientation can be inspected (Section~\ref{sec:interpretability}).

\begin{table*}[t]
\centering
\small
\caption{Ablation of the dual-stream gate and directional attention biases across five architectures. A: Unmodified backbone; B: A with dual stream-gate; C: B with windowed attention stage present, but the bias scalars are frozen at zero; D: C with WCAB/SCAB.}
\setlength{\tabcolsep}{3pt}
\renewcommand{\arraystretch}{1.3}
\label{tab:ablation}
\resizebox{\textwidth}{!}{%
\begin{tabular}{@{}ll cc cc cc cc@{}}
\toprule
& & \multicolumn{2}{c}{A: Base} & \multicolumn{2}{c}{B: {+}Gate} & \multicolumn{2}{c}{C: {+}Attn} & \multicolumn{2}{c}{D: {+}WCAB/SCAB } \\
\cmidrule(lr){3-4}\cmidrule(lr){5-6}\cmidrule(lr){7-8}\cmidrule(lr){9-10}
{Backbone} & {Params} & {F1} & {AUC-PR} & {F1} & {AUC-PR} & {F1} & {AUC-PR} & {F1} & {AUC-PR} \\
\midrule
SwinUNETR & 6.3M
  & {0.407 {\scriptsize[0.392, 0.420]}} & {0.334 {\scriptsize[0.321, 0.355]}}
  & {0.418 {\scriptsize[0.405, 0.430]}} & {0.358 {\scriptsize[0.344, 0.375]}}
  & {0.417 {\scriptsize[0.405, 0.430]}} & {0.359 {\scriptsize[0.342, 0.374]}}
  & {0.421 {\scriptsize[0.412, 0.436]}} & {0.364 {\scriptsize[0.350, 0.383]}} \\
MK-UNet & 0.32--0.40M
  & {0.411 {\scriptsize[0.400, 0.429]}} & {0.345 {\scriptsize[0.330, 0.367]}}
  & {0.413 {\scriptsize[0.398, 0.426]}} & {0.347 {\scriptsize[0.321, 0.357]}}
  & {0.406 {\scriptsize[0.388, 0.417]}} & {0.338 {\scriptsize[0.318, 0.353]}}
  & {0.415 {\scriptsize[0.398, 0.428]}} & {0.352 {\scriptsize[0.329, 0.367]}} \\
CAFIM-CNN & 3.0--4.3M
  & {0.416 {\scriptsize[0.402, 0.430]}} & {0.356 {\scriptsize[0.341, 0.374]}}
  & {0.417 {\scriptsize[0.402, 0.427]}} & {0.355 {\scriptsize[0.337, 0.369]}}
  & {0.414 {\scriptsize[0.404, 0.431]}} & {0.351 {\scriptsize[0.338, 0.372]}}
  & {0.414 {\scriptsize[0.402, 0.427]}} & {0.346 {\scriptsize[0.332, 0.362]}} \\
U-Net & 1.9--2.5M
  & {0.415 {\scriptsize[0.401, 0.426]}} & {0.346 {\scriptsize[0.327, 0.358]}}
  & {0.411 {\scriptsize[0.399, 0.422]}} & {0.352 {\scriptsize[0.337, 0.367]}}
  & {0.409 {\scriptsize[0.394, 0.422]}} & {0.344 {\scriptsize[0.324, 0.359]}}
  & {0.409 {\scriptsize[0.392, 0.420]}} & {0.346 {\scriptsize[0.319, 0.354]}} \\
Conv-AE & 1.7--2.3M
  & {0.396 {\scriptsize[0.382, 0.407]}} & {0.327 {\scriptsize[0.312, 0.344]}}
  & {0.395 {\scriptsize[0.380, 0.403]}} & {0.335 {\scriptsize[0.320, 0.348]}}
  & {0.390 {\scriptsize[0.379, 0.403]}} & {0.325 {\scriptsize[0.315, 0.344]}}
  & {0.391 {\scriptsize[0.381, 0.405]}} & {0.325 {\scriptsize[0.309, 0.341]}} \\
\bottomrule
\end{tabular}}
\end{table*}

To further assess these tool's behavior, the research conducts a feature ablation test and feature importance analysis on the SwinUNETR backbone (with the attention augmentation and dual-stream gate active). The rationale behind selecting one architecture was the fact that feature ablation assesses variable choice within the mechanism. Moreover, the classification metrics in Table~\ref{tab:ablation} were consistent across most models; however, the SwinUNETR backbone improved the most. Table~\ref{tab:varselect} examines if the model improvement is due to a purely statistical gain from adding informative features, or whether the directional components of wind and slope carry physical meaning within the model. This experiment replaces the wind and slope drivers with five alternate environmental variables: population density(pop/Population), Energy Release Component(ERC), Normalized Difference Vegetation Index(NDVI), Palmer Drought Severity Index(PDSI), and maximum temperature (Tempmax). These variables are assessed across seven configurations, holding all other training conditions fixed. Across the three seeds, no alternative differs significantly from the wind and slope reference. This pairing advantage is not seen from a classification view as two other pairings are nominally higher. Overall, the feature ablation shows that the wind and slope variables do not provide a distinctive accuracy improvement over other spatially varying signals. What the results do show is that the proposed attention-bias mechanism is robust to the choice of the driving variable. The case for selecting wind and slope (calculated from elevation) rests on field verifiability rather than on the best classification metrics. A bias keyed to observed wind and slope can be compared against conditions that can be measured in the field.

\par
\begin{table*}[t]
\centering
\small
\caption{Sensitivity of the SwinUNETR configuration D to the choice of bias variables. Each row replaces the wind driver(Bias 1) or the elevation driver (Bias 2) with alternative features.}
\setlength{\tabcolsep}{5pt}
\label{tab:varselect}
\resizebox{\textwidth}{!}{%
\begin{tabular}{lllccc}
\toprule
Experiment & Bias 1 & Bias 2 & F1 (mean $\pm$ std) & AUC-PR (mean $\pm$ std) & $\Delta$AUC-PR [95\% CI] \\
\midrule
\textbf{WCAB+SCAB} & \textbf{Wind} & \textbf{Elevation}
  & 0.4198 $\pm$ 0.0024 & 0.3594 $\pm$ 0.0012 & --- \\
\midrule
pop\_scab     & Wind       & Population & 0.4193 $\pm$ 0.0014 & 0.3596 $\pm$ 0.0030 & $+0.0001$ [$-0.0071$, $+0.0074$] \\
pop\_wcab     & Population & Elevation  & 0.4184 $\pm$ 0.0021 & 0.3601 $\pm$ 0.0043 & $+0.0007$ [$-0.0100$, $+0.0114$] \\
pop\_both     & Population & Population & 0.4206 $\pm$ 0.0015 & 0.3602 $\pm$ 0.0045 & $+0.0008$ [$-0.0108$, $+0.0124$] \\
erc\_scab$^a$ & Wind       & ERC        & 0.4084 $\pm$ 0.0146 & 0.3431 $\pm$ 0.0219 & $-0.0164$ [$-0.0860$, $+0.0533$] \\
ndvi\_scab    & Wind       & NDVI       & 0.4206 $\pm$ 0.0002 & 0.3614 $\pm$ 0.0016 & $+0.0020$ [$-0.0018$, $+0.0057$] \\
pdsi\_scab    & Wind       & PDSI       & 0.4181 $\pm$ 0.0019 & 0.3593 $\pm$ 0.0014 & $-0.0001$ [$-0.0042$, $+0.0040$] \\
tempmax\_scab & Wind       & Temp.\ max & 0.4179 $\pm$ 0.0012 & 0.3584 $\pm$ 0.0049 & $-0.0010$ [$-0.0139$, $+0.0118$] \\
\bottomrule
\end{tabular}}
\end{table*}

One reason the alternatives in Table~\ref{tab:varselect} perform similarly is the dominance of the previous-day fire mask. Permuting this channel costs $\Delta\mathrm{F1} = -0.25$, an order of magnitude more than any other input. Since every configuration shares the previous-day fire mask channel, the marginal gain available to any second bias variable is compressed. Most of the predictive signal is already carried by prior-day fire geometry. This observation also highlights the role of data resolution. The fire masks are derived from the MODIS MOD14A1 daily thermal-anomaly product at $1\,\mathrm{km}$ (Section~\ref{sec:data}), whereas the wind fields originate from GRIDMET at $\sim$4\,$\mathrm{km}$ resolution and vary little within a single attention window at the scales considered here. The elevation channel is aggregated from $\sim$30\,$\mathrm{m}$ SRTM data to the same $1\,\mathrm{km}$ grid. The wind bias is therefore driven by an input roughly four times coarser than the fire mask it competes with, and the slope bias by terrain aggregated to the same grid as the fire mask.

\begin{figure*}[t]
\centering
\includegraphics[width=\textwidth]{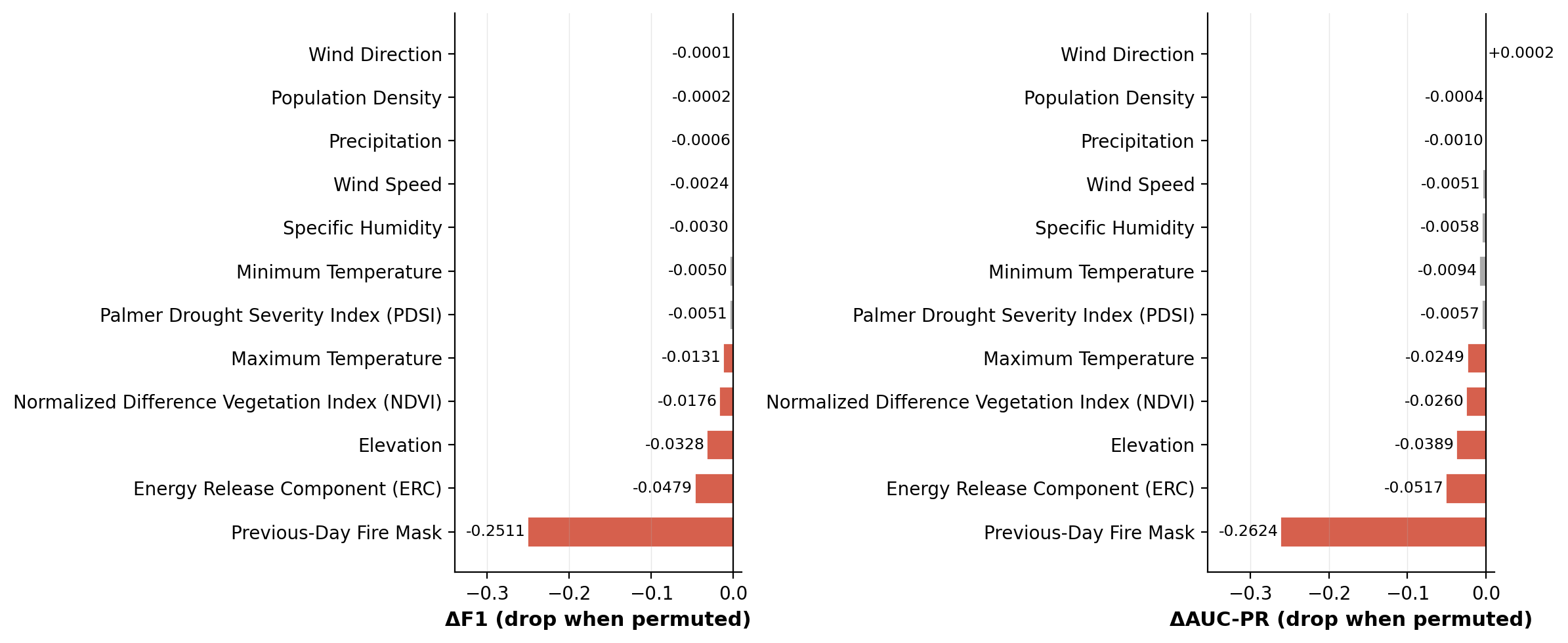}
\caption{Permutation feature importance for the SwinUNETR with all proposed methods. The previous-day fire mask has a clear feature importance as F1 and AUC-PR drop rather extensively. }
\label{fig:feature_importance}
\end{figure*}

\subsection{Directional Bias Audit}
\label{sec:interpretability}
To explore the directional bias further, the research tests how faithfully these biases reflect the observed wind, the terrain gradient, and the fire's next-day spread. For all five architectures, the learned attention-bias field is extracted at its most physics-converged block as seen in Table \ref{tab:block_alignment}. The bias field's per-pixel attention-received map is reduced to a single direction by taking the centroid of its positive part relative to the tile center. This direction is then compared, via the angle between two vectors, against the wind read directly from an input channel, the tile-mean elevation gradient for slope, or the empirical displacement between the prior and next-day fire perimeter centroids for spread. The angular deviation between the field's centroid and observed wind direction is measured across the test set for every architecture. When examining wind alignment, each architecture's bias field levels off near the same ceiling, with roughly 98.3--98.6\% of test tiles falling within 45\textdegree{} of the observed wind for all five models. The supplementary material section shows that several blocks per architecture independently converge to the same finding. This indicates that the audit is not an artifact of selecting a single best-performing block.

When examining next-day spread, the same field shows no reliable correspondence with where the fire actually spreads the following day (28.6--29.0\% within $45^{\circ}$, barely above the $\sim$25\% expected by chance), and this null test result replicates identically across all models. The alignment audit verifies the orientation of the prescribed bias field. It does not establish that the complete attention distribution or predicted fire movement follows the same direction. The limited correspondence with observed next-day displacement therefore bounds the physical interpretation of this audit. The physical grounding that makes the bias traceable does not drive the model's accuracy, which is instead dominated by the previous-day fire mask, as the permutation analysis above shows ($\Delta\mathrm{F1} = -0.25$). Extending the same audit to the slope bias shows that it does not have similar behavior to the wind bias. This could be because the wind channel is spatially constant across the $64\times64$ tile, unlike elevation gradient that drives the slope bias, varies pixel by pixel within a tile. Alignment with the tile-mean upslope direction is modest across the multi-window block (20.8--48\%, Table~\ref{tab:block_alignment}). This is only slightly above the $\sim$25\% chance floor. This asymmetry follows largely from the spatial structure of the two inputs rather than from the mechanism itself. When wind is nearly uniform within a tile, any window-level average recovers the tile-level value by construction. This makes its alignment check close to a direct readout of a single scalar sign. However, the slope varies across a tile and alignment is based on complex geometry. Consistent with this, the model that recovers wind-like slope alignment, where a single window spans the entire tile, is the SwinUNETR because slope collapses back into the same spatially uniform regime as wind. Therefore, the architectural methodology of window sliding plays a role when assessing directional bias.

\begin{figure*}[h!]
\centering
\includegraphics[width=\textwidth,height=0.6\textheight,keepaspectratio]{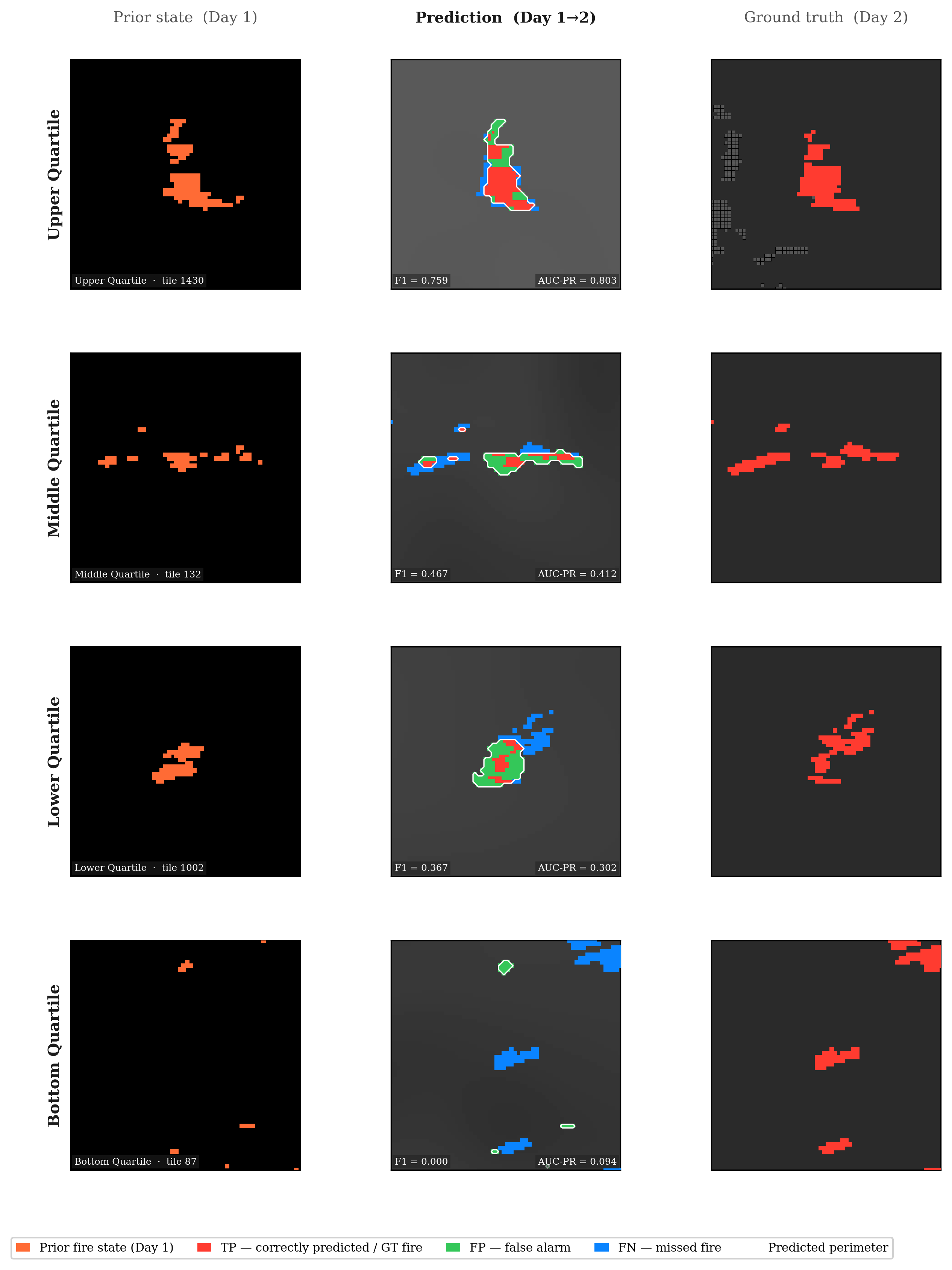}
\caption{Representative test-tile predictions from gate and attention augmentation components. 
SwinUNETR configuration, spanning the upper (top 25\%), middle,
lower (bottom 25\%), and bottom (worst performing tile) quartiles of per-tile F1 on the Huot et al. (2022) benchmark test
set. Each row shows the prior-day fire state, the model's Day
1$\to$2 prediction (TP/FP/FN relative to ground truth), and the
observed Day 2 fire extent.}
\label{fig:TwoMechMap}
\end{figure*}

Overall, the bias field produces a stable readout of the directional preference imposed on the attention scores. Its orientation matches the input wind direction consistently, so a forecaster can verify that the prescribed prior was applied as intended. This does not verify the model's reasoning as a whole, and the prior does not by itself forecast the next-day spread of the fire.  However, the slope bias has a smaller converged $\gamma$ magnitude. Slope encodes a spatially finer-grained signal that a single global direction cannot be fully characterized. Therefore, a practitioner should treat the wind check as the primary tile-level verification and slope's contribution as a secondary check. A forecaster can use the wind field in two ways: (1) to confirm that the prescribed wind prior was applied in expected orientation (even when the prediction is wrong), or (2) to flag a case where the two biases visibly disagree. Because this behavior holds across the five architectures, the mechanism offers an inspectable directional prior at a small and host-dependent cost in classification performance (Table ~\ref{tab:ablation}).  

\par
\begin{table}[t]
\centering
\small
\caption{Directional alignment of the learned bias field. Percentage of test tiles whose bias field centroid direction lies within $45^{\circ}$ of the input wind direction (Wind), the tile-mean upslope direction (Slope), and the observed displacement between the prior- and next-day fire centroids (Spread).}
\setlength{\tabcolsep}{4pt}
\label{tab:block_alignment}
\begin{tabular}{lccc}
\toprule
Architecture & Wind (\%) & Slope (\%) & Spread (\%) \\
\midrule
SwinUNETR  & 98.5 & 47.8             & 28.6 \\
MK-UNet    & 98.3 & 37.7            & 28.8 \\
CAFIM-CNN  & 98.6 & 34.3            & 28.7 \\
Plain U-Net & 98.5 & 48.0           & 28.6 \\
Conv-AE    & 98.5 & 20.8            & 29.0 \\
\bottomrule
\end{tabular}
\end{table}

To complement the results presented in Table~\ref{tab:ablation}, Figure~\ref{fig:TwoMechMap} shows predictions, on the test set, from the SwinUNETR model with the gating and attention augmentation mechanisms added. This figure lines up with the feature ablation. The previous-day fire mask is the dominant cue, so the model is strongest when the fire perimeters are established and weakest when they are sparse. Despite the previous fire mask being a dominate feature Table \ref{tab:block_alignment} show how the wind components influence the model.  The best tile (F1 = 0.759) shows a well-localized boundary on an established, contiguous fire. The false negatives concentrate around the fire perimeter, while false positives fall within the fire boundary. This trend holds in the Middel (F1 = 0.467) and Lower-quartile (F1 = 0.367) cases as well. Together, these indicate that the model favors persistence over spread. It confidently carries the existing fire forward — occasionally over-predicting interior pixels that in fact extinguish. Conversely, the advancing perimeter, where next-day growth actually occurs, remains the harder prediction target. Future improvements should therefore focus on the missed cases along the fire perimeter. The worst case (F1 = 0.000) corresponds to a sparse, fragmented prior-day fire with minimal existing perimeter, and both small fronts are missed entirely. Given that the previous-day fire mask is by far the dominant predictive feature on this benchmark, tiles that offer almost no prior perimeter leave the model without its primary anchor. This failure mode reflects the extreme end of the domain's class imbalance rather than a mechanism-specific shortcoming.

\subsection{Physics Feature Retrieval Augmented Output Correction Results}
\label{sec:rag_results}
The third component is the retrieval correction. The rationale for including this component is to make the correction traceable to specific retrieved training tiles that can be inspected alongside the forecast.  Section~\ref{sec:interpretability} established that the directional attention priors are geometrically coherent and consistent across architectures; however, their accuracy advantage over non-physical spatial controls is small. Moreover, the physics evidence for the first two mechanisms is structural rather than behavioral. They show the attention augmentation priors were built into the computation and do not demonstrate that the model's predictions depend on the physical content of the wind and slope inputs. The PFRAOC mechanism admits a different form of evidence. This is because it consults physical descriptors at inference rather than encoding them once during training. Therefore, its dependence on those descriptors can be probed directly. This is achieved by corrupting the physical knowledge of the features in an unchanged trained model and observing whether performance degrades. The knowledge base, built as described in Section~\ref{sec:rag}, is queried at inference for the $K{=}3$ nearest training tiles. Their observed next-day spread masks are averaged and passed through a lightweight correction network added to the frozen base model's logits.

Table~\ref{tab:rag_transfer} shows both configurations (correct physics and wrong physics) of the retrieval mechanism across five architectures.

The individual classification metrics are reported as a three-seed average, while the confidence intervals reflect tile-resampling uncertainty of a single seed. Seed-to-seed variability is reported separately as the standard deviation. The SwinUNETR retrieval raises the mean test AUC-PR from 0.3640 to 0.3673 across three seeds. The gain is modest and its bootstrap confidence interval overlaps the two-mechanism model's. By contrast, the MK-UNet and CAFIM-CNN F1 scores at the frozen threshold decline, while AUC-PR rises. Therefore, the retrieval benefit is metric specific. AUC-PR will be examined when evaluating the correct vs. wrong physics ablation. A similar pattern emerges between the two-mechanism (Base) and three-mechanism (PFRAOC) configurations on the MK-UNet ($\Delta$AUC-PR $= +0.0053$) and CAFIM-CNN ($\Delta$AUC-PR $= +0.0037$). Under the wrong-physics retrieval, AUC-PR is lower than under correct retrieval in every host and every seed tested. The SwinUNETR wrong-physics configuration erases nearly the entire retrieval benefit, while on MK-UNet and CAFIM-CNN the reduction is partial. Threshold F1 under wrong-physics retrieval returns to the level of the uncorrected base model on every host. Corrupting the descriptor removes the correct configuration's AUC-PR gain. The pooled confidence intervals for correct and wrong-physics retrieval overlap in all cases. The ablation is therefore reported as a consistent directional observation across independently trained models rather than as a statistically established effect, and the differences should be read as estimates with uncertainty. Changing the wind components of the retrieval query reduced mean AUC-PR relative to the unmodified query across the tested hosts. This indicates sensitivity to the retrieval descriptor, but does not rule out dependence on related incidents or establish generalization to a chronologically independent bank.

The research tested whether the analog prior could be moved into training instead of inference. The retrieved prior enters as a consistency loss (weight $\lambda_r$), and the deployed model becomes a single forward pass with no retrieval index. This was assessed on SwinUNETR, and its best setting ($\lambda_r = 0.1$) reaches AUC-PR 0.3608 [0.3444, 0.3759]. This is statistically indistinguishable from the post-hoc design. The in-loop design offers no accuracy advantage while forfeiting the inspectable retrieved tiles. Specifically, including the prior into the weights discards the inspectable analogs and freezes the knowledge base at training time. Conversely, a post-hoc bank can be extended with each new fire season without retraining. This experiment supports the post-hoc design rather than replacing it. The Supplementary Material section (Fig. ~\ref{fig:vuln_rag}) illustrates the retrieved tiles for representative test cases.

The value of this mechanism does not rest on the aggregate accuracy gain reported in Table~\ref{tab:rag_transfer}. That gain is modest and statistically inconclusive from the pooled intervals alone. Rather, its practical contribution is that each correction is traceable to specific retrieved training tiles that can be inspected alongside the forecast, at a small cost in threshold F1 on some hosts (Table ~\ref{tab:rag_transfer}).  Thus, a forecaster can judge not only the model's prediction but also the likeness of the ongoing event to the retrieved tiles, a comparison that generic model outputs do not permit. Future research can assess whether this comparison improves analysts' judgment as discussed in Section \ref{sec:discussion}. 

The main limitation of the retrieval mechanism is that the bank is built from the training and validation distribution. Because the released tensors carry no event or date identifiers, a chronologically held-out bank was not constructed here. Recovering or reconstructing such identifiers, or evaluating against an independently sourced fire dataset, is a prerequisite for that test. The wrong-physics ablation, which degrades performance while holding the bank fixed, establishes sensitivity to the retrieval query but does not rule out memorization of related tiles. To probe this further, the supplementary analysis shows the results for five control experiments. These are, a random-neighbor retrieval, a fixed dataset-mean mask, a train-only bank, and two retrieval-pool restrictions to fire-containing bank tiles.

\begin{table*}[t]
\centering
\small
\caption{The Physics Feature Retrieval Augmented Output Correction mechanism implemented across the five hosts on the test set. Base: configuration D of Table \ref{tab:ablation} is used to compare how each model performs when all three mechanisms are added to each model (+PFRAOC(correct)). }
\setlength{\tabcolsep}{4pt}
\label{tab:rag_transfer}
\begin{tabular}{llcc}
\toprule
Host & Configuration & F1 (mean $\pm$ std [CI]) & AUC-PR (mean $\pm$ std [CI]) \\
\midrule
\multirow{3}{*}{SwinUNETR}
 & Base                 & 0.4209 $\pm$ 0.0039 [0.4116, 0.4364] & 0.3640 $\pm$ 0.0028 [0.3495, 0.3833] \\
 & +PFRAOC (correct)        & 0.4216 $\pm$ 0.0028 [0.4099, 0.4372]           & 0.3673 $\pm$ 0.0033 [0.3534, 0.3871]           \\
 & +PFRAOC (wrong physics)  & 0.4227 $\pm$ 0.0021 [0.4121, 0.4366]  & 0.3638 $\pm$ 0.0026 [0.3506, 0.3839] \\
          
\midrule
\multirow{3}{*}{MK-UNet}
 & Base                 & 0.4151 $\pm$ 0.0038 [0.3981, 0.4277]  & 0.3520 $\pm$ 0.0033 [0.3293, 0.3674]  \\
 & +PFRAOC (correct)        & 0.4105 $\pm$ 0.0028 [0.3967, 0.4258]  & 0.3573 $\pm$ 0.0019 [0.3372, 0.3734]  \\
 & +PFRAOC (wrong physics)  & 0.4162 $\pm$ 0.0026 [0.4021, 0.4293]  & 0.3535 $\pm$ 0.0018 [0.3339, 0.3692]  \\
\midrule
\multirow{3}{*}{CAFIM-CNN}
 & Base                 & 0.4139 $\pm$ 0.0031 [0.4023, 0.4268]  & 0.3460 $\pm$ 0.0035 [0.3318, 0.3615]  \\
 & +PFRAOC (correct)        & 0.4054 $\pm$ 0.0059 [0.3902, 0.4166]  & 0.3497 $\pm$ 0.0092 [0.3320, 0.3614]  \\
 & +PFRAOC (wrong physics)  & 0.4132 $\pm$ 0.0045 [0.4001, 0.4227]  & 0.3449 $\pm$ 0.0082 [0.3270, 0.3547]  \\
\midrule
\multirow{3}{*}{U-Net}
 & Base                 & 0.4090 $\pm$ 0.0032 [0.3919, 0.4197]  & 0.3456 $\pm$ 0.0089 [0.3185, 0.3539]  \\
 & +PFRAOC (correct)        & 0.4095 $\pm$ 0.0049 [0.3900, 0.4190]  & 0.3515 $\pm$ 0.0074 [0.3262, 0.3609]  \\
 & +PFRAOC (wrong physics)  & 0.4112 $\pm$ 0.0020 [0.3944, 0.4216]  & 0.3473 $\pm$ 0.0064 [0.3226, 0.3559]  \\
\midrule
\multirow{3}{*}{Conv-AE}
 & Base                 & 0.3913 $\pm$ 0.0027 [0.3806, 0.4048]  & 0.3254 $\pm$ 0.0009 [0.3093, 0.3411]  \\
 & +PFRAOC (correct)        & 0.3904 $\pm$ 0.0026 [0.3807, 0.4048]  & 0.3305 $\pm$ 0.0044 [0.3093, 0.3412]  \\
 & +PFRAOC (wrong physics)  & 0.3925 $\pm$ 0.0020 [0.3807, 0.4048]  & 0.3280 $\pm$ 0.0017 [0.3093, 0.3412]  \\
\bottomrule
\end{tabular}
\\[2pt]

\end{table*}

\begin{figure}[t]
    \centering
    \includegraphics[width=\columnwidth]{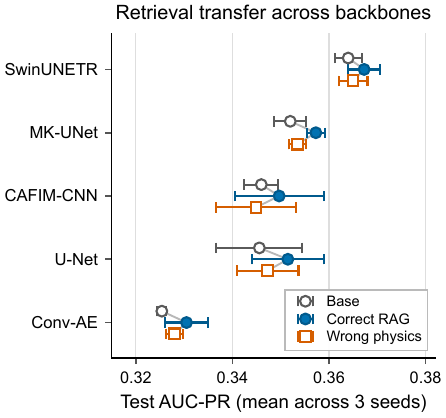}
    \caption{Retrieval transfer across the five backbones in Table~\ref{tab:rag_transfer}. Markers and horizontal bars show the mean and one standard deviation over three seeds. Lines link the base model, correct-physics retrieval, and wrong-physics retrieval within each host. Correct-physics retrieval has higher mean AUC-PR than wrong-physics retrieval for every backbone.}
    \label{fig:rag_transfer}
\end{figure}

\subsection{Ensemble Configuration Results}
\label{sec:ens_results}
The research evaluated the three proposed modules on individual architectures. To further demonstrate the flexibility they provide, the research now evaluates the performance and behavior of the mechanisms when multiple models are used at once for predictions. Each model independently produces a full $64\times64$ map of the probability that each pixel is active fire on day $t{+}1$. Then those probabilities are averaged together based on the number of models. The first configuration (config1), all five models with the three mechanisms are tested in an ensemble. This resulted in an AUC-PR of 0.3718 and an F1 score of 0.4246. Then, the U-Net and Conv-AE models are omitted from the ensemble. The three remaining models, with all modules, are assessed for the second configuration (config2). This resulted in an AUC-PR of 0.3729  and an F1 score of 0.4267. As seen, merely combining model predictions is not an avenue for classification enhancement. The U-Net and Conv-AE models contribute little of their own since they gain almost no benefit from the mechanisms individually, as seen in Table~\ref{tab:ablation}. All models in the first two ensemble configurations had every proposed enhancement. Next, the three-architecture configuration is assessed without any mechanism for the third configuration (config3). This resulted in an AUC-PR of 0.3673 and F1 of 0.4245. This  configuration obtained the lowest classification metrics among all ensembles. A single SwinUNETR with all three mechanisms reaches a comparable AUC-PR (three-seed mean 0.3673, Table ~\ref{tab:rag_transfer}). As seen, architectural diversity without the mechanisms did not outperform a single well-matched host.

When examining individual architectures, the CAFIM-CNN performed worse with all the proposed components. This is consistent with the architecture comparison of Han et al. (2026), which found that transformer-style components degrade performance on their model \cite{han2026}. Moreover, its cross-modal fusion module is already designed to model fuel--weather interactions. Given this information, the final ensemble configuration (config4) included the CAFIM-CNN without any of this research's enhancements alongside the SwinUNETR and MK-UNet models with all proposed elements. This achieved the highest mean scores among all configurations across the three seeds (AUC-PR 0.3790, F1 0.4292). This demonstrates the flexibility the mechanisms have when experimenting with different deep learning architectures. As more models are developed, these design elements show they can be a valuable add-on to existing models. This could be an individual model or in conjunction with other models, and the value added does not dilute architectures that already perform well without them.

Although ensembling increases the complexity of the system, the inspection objects of the augmented members remain available because the combination is a fixed, unweighted mean over the members' probability maps. Nothing is learned in the fusion step, so every ensemble prediction decomposes exactly into its members' contributions. Equal-weight averaging makes the aggregate decomposable into its members. It does not make every member's prediction physically interpretable. The decomposition is illustrated in the Supplementary Material (Fig.~\ref{fig:walkthrough_ensemble})

\begin{table*}[t]
\centering
\small
\caption{Inference-time deployment results for four ensemble
configurations. Config.~1 averages all five architectures, each with the
gating and attention-bias mechanisms; Config.~2 averages SwinUNETR,
MK-UNet, and CAFIM-CNN with the same mechanism configuration; Config.~3
averages the same three architectures without any mechanisms; Config.~4
combines SwinUNETR and MK-UNet equipped with all three mechanisms
(including the retrieval correction) with a vanilla CAFIM-CNN. The single
retrieval-corrected SwinUNETR model is shown for reference. Values are
means $\pm$ standard deviation over three seeds; brackets are 95\%
tile-bootstrap confidence intervals.}
\setlength{\tabcolsep}{4pt}
\label{tab:ensembles}
\begin{tabular}{lccc}
\toprule
Configuration & Members & F1 (mean $\pm$ std [CI]) & AUC-PR (mean $\pm$ std [CI]) \\
\midrule
Single model (SwinUNETR, all mechanisms) & 1 & 0.4216 $\pm$ 0.0028 [0.4099, 0.4372] & 0.3673 $\pm$ 0.0033 [0.3534, 0.3871] \\
\midrule
Config.~1 & 5 & 0.4246 $\pm$ 0.0005 [0.4107, 0.4368] & 0.3718 $\pm$ 0.0005 [0.3545, 0.3859] \\
Config.~2 & 3 & 0.4267 $\pm$ 0.0003 [0.4132, 0.4390] & 0.3729 $\pm$ 0.0015 [0.3565, 0.3889] \\
Config.~3 & 3 & 0.4245 $\pm$ 0.0011 [0.4088, 0.4368] & 0.3673 $\pm$ 0.0019 [0.3523, 0.3861] \\
Config.~4 & 3 & 0.4292 $\pm$ 0.0016 [0.4148, 0.4411] & 0.3790 $\pm$ 0.0025 [0.3632, 0.3974] \\
\bottomrule
\end{tabular}
\end{table*}

\subsection{Calibration and Cost}
\label{sec:cal_cost_results}
To complement the classification performance, the research explores the computational cost (parameters, single-tile latency, and peak memory), and probability calibration (expected calibration error and Brier score) for all five architectures and the four ensemble configurations. 

Table ~\ref{tab:cost} shows the parameter count, single-tile GPU latency, and peak memory batch 32, for all experiments. The Mk-UNet configurations have the fewest parameters because its multi-kernel design routes computation through parallel depthwise branches rather than dense convolutions. Moreover, the spatial and channel-wise filtering  yields fewer weights per layer compared to dense fusion convolutions such as those seen in CAFIM-CNN or attention projections in SwinUNETR. The Conv-AE has the lowest latency because it lacks skip connections, so the forward pass carries no feature-map concatenation overhead. It is purely a convolutional backbone with no attention windowing. The CAFIM-CNN follows a similar pattern despite having roughly ten times more parameters than MK-UNet, since dense convolutions are more efficiently parallelized on GPU hardware than many small windowed attention operations of MK-UNet and SwinUNETR. The SwinUNETR has the highest computational cost among individual models and the highest individual classification scores. This trade-off may be justified when a single model's classification performance and inspection outputs are valued over per-inference cost. Where compute is constrained, MK-UNet supports all three mechanisms at a small fraction of the cost with somewhat lower classification scores.

Ensemble costs are reported once per configuration as the sum of the member costs. The parameter counts exclude the retrieval bank, and the summed latency and peak memory assume sequential execution of the members, so they are approximations rather than end-to-end measurements. On this basis the mixed ensemble is the second cheapest configuration, close to the all-vanilla ensemble (9.72M/32.5ms/688.3MB vs 9.63M/23.94ms/641.3MB). The only overhead is the mechanisms on the two architectures that benefit them. Overall, the proposed additions increase computational cost in exchange for the inspection outputs described throughout the manuscript, and the range of configurations gives users flexibility depending on their computational resource.  

Calibration follows a more architecture-dependent pattern than cost, and the changes are mixed (Table ~\ref{tab:calibration}. For SwinUNETR, ECE does not improve from the vanilla model (0.0297) to configuration D (0.0306); only the retrieval-corrected configuration attains lower ECE and Brier score (0.0281 and 0.01429). For MK-UNet, the directional biases lower both ECE and Brier scores relative to the baseline. Moreover, retrieval lowers ECE further while slightly worsening the Brier score (0.01476 to 0.01481). For CAFIM-CNN, U-Net, and Conv-AE, the vanilla configuration has the lowest ECE and Brier score. Among the ensembles, the mixed configuration has the lowest ECE and Brier score (0.0271 and 0.01327), which is consistent with combining each architecture's own better-calibrated configuration. However, the vanilla CAFIM-CNN alone (0.0230 and 0.01295) remains better calibrated than any ensemble.

The cost and calibration results describe the trade-offs of integrating the mechanisms in each architecture. The mixed ensemble combines the highest mean classification scores in this study with competitive calibration at roughly the cost of three single models. Where compute is limited, MK-UNet with all mechanism provide the inspection outputs at one of the smallest parameter counts. Additionally, its calibration is close to that of the mixed ensemble and its classification scores are somewhat below it. Individually, the SwinUNETR with all three mechanisms has the highest classification scores and the best calibration among the augmented single models. The reported ECE and Brier scores describe retrospective probability performance under this evaluation. Their mixed changes across configurations do not establish calibration in future incidents or suitability for direct operational decision thresholds, and reference values (an all-zero forecast has a Brier score equal to the test prevalence $p$, and a constant-prevalence forecast $p(1-p)$), reliability diagrams with bin counts, and persistence baselines would be needed to interpret the absolute values.

\par
\begin{table*}[h!]
\centering
\caption{Expected calibration error (ECE) and Brier score for each architecture configurations A-D of Table \ref{tab:ablation} and with the post-hoc retrieval correction (D+PFRAOC). The four ensemble configurations are also shown. }
\footnotesize
\setlength{\tabcolsep}{2.5pt}
\renewcommand{\arraystretch}{1.12}
\label{tab:calibration}
\begin{tabularx}{\textwidth}{@{}>{\raggedright\arraybackslash}X
>{\centering\arraybackslash}m{1.15cm}
>{\centering\arraybackslash}m{1.25cm}
>{\centering\arraybackslash}m{1.50cm}
>{\centering\arraybackslash}m{1.80cm}
>{\centering\arraybackslash}m{1.40cm}@{}}
\toprule
\textbf{Architecture} & \shortstack{A\\(base)} & \shortstack{B\\(+gate)} & \shortstack{C\\(+attn$^\dagger$)} & \shortstack{D\\(+WCAB/\\SCAB)} & \shortstack{D+\\PFRAOC} \\
\midrule
SwinUNETR ECE   & 0.0297 & 0.0325 & 0.0307 & 0.0306 & 0.0281 \\
SwinUNETR Brier & 0.01524 & 0.01546 & 0.01509 & 0.01501 & 0.01429 \\
\midrule
MK-UNet ECE     & 0.0295 & 0.0292 & 0.0290 & 0.0284 & 0.0277 \\
MK-UNet Brier   & 0.01519 & 0.01527 & 0.01532 & 0.01476 & 0.01481 \\
\midrule
CAFIM-CNN ECE   & 0.0230 & 0.0299 & 0.0293 & 0.0272 & 0.0285 \\
CAFIM-CNN Brier & 0.01295 & 0.01385 & 0.01431 & 0.01334 & 0.01468 \\
\midrule
Plain U-Net ECE   & 0.0292 & 0.0337 & 0.0301 & 0.0311 & 0.0312 \\
Plain U-Net Brier & 0.01490 & 0.01571 & 0.01507 & 0.01542 & 0.01579 \\
\midrule
Conv-AE ECE       & 0.0289 & 0.0344 & 0.0307 & 0.0299 & 0.0300 \\
Conv-AE Brier     & 0.01504 & 0.01628 & 0.01552 & 0.01515 & 0.01516 \\
\midrule
\multicolumn{6}{@{}l@{}}{\textit{Ensemble configurations (combined probability map, seed 42)}} \\
\midrule
5 arch., full mechanism (5-member) ECE   & \multicolumn{5}{c@{}}{0.0308} \\
5 arch., full mechanism (5-member) Brier & \multicolumn{5}{c@{}}{0.01399} \\
\addlinespace
3 arch., full mechanism (3-member) ECE    & \multicolumn{5}{c@{}}{0.0310} \\
3 arch., full mechanism (3-member) Brier  & \multicolumn{5}{c@{}}{0.01391} \\
\addlinespace
3 arch., no mechanisms (vanilla) ECE      & \multicolumn{5}{c@{}}{0.0272} \\
3 arch., no mechanisms (vanilla) Brier    & \multicolumn{5}{c@{}}{0.01340} \\
\addlinespace
Mixed: Swin+MK mech., CAFIM-CNN vanilla, ECE   & \multicolumn{5}{c@{}}{0.0271} \\
Mixed: Swin+MK mech., CAFIM-CNN vanilla, Brier & \multicolumn{5}{c@{}}{0.01327} \\
\bottomrule
\end{tabularx}
\end{table*}

% % ============================================================
\par
\begin{table*}[t!]
\centering
\setlength{\tabcolsep}{3pt}
\footnotesize
\renewcommand{\arraystretch}{1.08}
\caption{Computational cost at inference. The parameter count, median single tile GPU latency (batch size 1), and peak GPU memory (batch size 32) for the individual and ensemble architectures.}
\label{tab:cost}
\begin{tabularx}{\textwidth}{@{}>{\raggedright\arraybackslash}Xrrr@{}}
\toprule
Configuration & Params & Latency (ms, $b{=}1$) & Peak mem (MB, $b{=}32$) \\
\midrule
\multicolumn{4}{l}{\textit{Individual architectures}} \\
SwinUNETR --- vanilla (A)                  & 6.297M & 13.42 & 158.5 \\
SwinUNETR --- full mechanism (D)           & 6.297M & 15.73 & 186.3 \\
SwinUNETR --- full mechanism + RAG (D+PFRAOC) & 6.299M & 16.27 & 187.8 \\
MK-UNet --- vanilla (A)                    & 0.317M &  7.35 & 129.6 \\
MK-UNet --- full mechanism (D)             & 0.402M & 12.53 & 146.2 \\
MK-UNet --- full mechanism + RAG (D+PFRAOC)   & 0.404M & 13.09 & 147.3 \\
CAFIM-CNN --- vanilla (A)                  & 3.019M &  3.17 & 353.2 \\
CAFIM-CNN --- full mechanism (D)           & 4.341M &  8.25 & 369.6 \\
CAFIM-CNN --- full mechanism + RAG (D+PFRAOC) & 4.343M &  8.80 & 371.7 \\
Plain U-Net --- vanilla (A)                & 1.928M &  1.12 & 125.5 \\
Plain U-Net --- full mechanism (D)         & 2.460M &  7.70 & 152.4 \\
Conv-AE --- vanilla (A)                    & 1.735M &  1.08 & 107.9 \\
Conv-AE --- full mechanism (D)             & 2.266M &  5.83 & 134.6 \\
\midrule
\multicolumn{4}{l}{\textit{Ensembles (3-seed mean classification; cost is seed-independent)}} \\
5-arch, full mechanism (5 members)        & 15.77M & 50.04 & 989.1 \\
3-arch, full mechanism (3 members)         & 11.04M & 36.51 & 702.1 \\
3-arch, all vanilla (control)              &  9.63M & 23.94 & 641.3 \\
{3-arch, mixed} (Swin+MK full mech.\ + PFRAOC, CAFIM-CNN vanilla) & {9.72M} & {32.53} & {688.3} \\
\bottomrule
\end{tabularx}
\end{table*}

\clearpage
\begin{figure}[!t]
    \centering
    \includegraphics[width=\columnwidth]
        {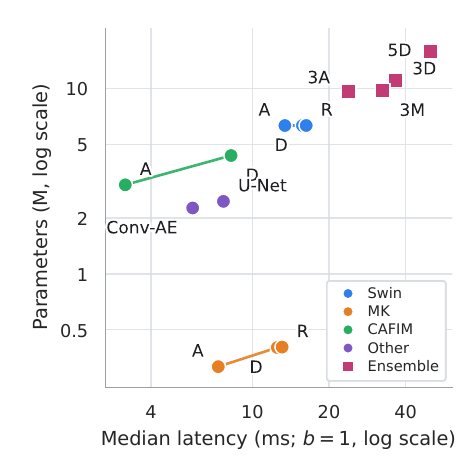}
    \caption{
    Compute-cost trade-off across individual architectures and ensembles.
    Horizontal and vertical positions report median batch-1 latency and
    parameter count, respectively; both axes use logarithmic scales.
    Circles denote individual architectures and squares denote ensembles.
    Within each backbone, connecting lines show the progression from vanilla
    (\textnormal{A}) to the full mechanism
    (\textnormal{D}) and to the full mechanism with PFRAOC
    (\textnormal{R}). Ensemble abbreviations denote the five-architecture
    full-mechanism ensemble (\textnormal{5D}), three-architecture
    full-mechanism ensemble (\textnormal{3D}), all-vanilla control
    (\textnormal{3A}), and mixed ensemble (\textnormal{3M}).
    }
    \label{fig:compute_cost_tradeoff}
\end{figure}

\section{Discussion}
\label{sec:discussion}
The principal finding of this study is that explicit directional priors and retrieved examples can be integrated into different wildfire prediction architectures, while their predictive effects remain conditional on the host. The SwinUNETR model was able to take advantage of all three mechanisms despite the gating component being the least transferable module. The retrieval increased mean AUC-PR relative to the corresponding augmented model across all five hosts, but it did not consistently improve F1 (Table \ref{tab:rag_transfer}). The directional audit likewise separates inspectability from predictive alignment. Specifically, the bias field closely followed the input wind, while its alignment with observed next-day fire displacement was much weaker (Table \ref{tab:block_alignment}). These results support treating the modules as identifiable components whose behavior can be examined, rather than as evidence that the complete model has acquired robust physical reasoning. The mixed ensemble obtained the strongest reported aggregate predictive scores, illustrating the value of selective module use. Although, the observed differences do not establish statistical superiority over the alternative configurations evaluated in this study. 

Several host-dependent patterns prompt the need for mechanistic explanations. The present experiments focus on demonstrating how these modules can be utilized and interpreted rather than assessing how they alter or take advantage of specific architectural behavior. Therefore, the following discussion offers hypotheses as to how the proposed methods interact with the tested models. The attention bias produced its clearest gains on SwinUNETR, where attention is native, and on MK-UNet, whose multi-kernel design arbitrates among parallel receptive-field branches; on U-Net and Conv-AE, the grafted attention stage may be an unconstrained pathway that a small, label-sparse dataset does not supply enough signal to exploit; and CAFIM-CNN's built-in cross-modal fusion may already encode the fuel--weather interactions the prior supplies. The gate showed the opposite pattern, contributing mainly on SwinUNETR, which has no native means of prioritizing spatial regions, whereas MK-UNet and CAFIM-CNN perform their own spatial arbitration internally. Retrieval raised mean AUC-PR on all five hosts, including U-Net ($+0.0059$) and Conv-AE ($+0.0051$), whose attention-bias gains were negligible; a plausible reason is that the correction network is trained against a frozen backbone and so bypasses the joint-training capacity constraint. Self-attention and multiscale convolutions already provide forms of spatial processing, so these explanations require targeted experiments before they can be treated as established.

Related concerns have been raised in the deep learning literature. Rudin (2019) argues that high-stakes decisions should not rely on black-box models~\cite{rudin2019}. Doshi-Velez and Kim~\cite{doshi2017towards} and Lipton~\cite{lipton2018mythos} argue that interpretability claims must be evaluated as measurable properties rather than asserted qualitatively, and Turb\'{e} et al.~\cite{turbe2023evaluation} show that post-hoc attribution methods disagree with one another under systematic evaluation, undermining their use as a reliable audit trail. In the wildfire domain specifically, Becker et al.\ (2026) found that their most physically consistent models were also their least accurate~\cite{becker2026assessing} and advocated closing that gap rather than treating accuracy and physical consistency as unavoidably opposed. The approach taken here responds to that call in a bounded way: rather than layering post-hoc explanation onto a black box or adopting a weaker transparent model, it embeds identifiable components inside several established architectures. The evidence presented supports the inspectability of those components and documents their predictive effects; it does not show that the complete models have become interpretable, and accuracy is not preserved on every host (Tables~\ref{tab:ablation} and~\ref{tab:rag_transfer}).

This study's scientific contribution is a modular approach to examining how explicit domain cues interact with learned representations in next-day wildfire prediction. Wind and slope enter attention as identifiable additive terms, while a separate correction module uses retrieved next-day fire masks selected through an interpretable environmental descriptor. A staged ablation distinguishes the effects of fire-conditioned gating, added attention capacity, directional bias, and retrieval correction. The comparison across five backbones shows that integration is feasible across different model designs, while predictive benefits depend on the host and the evaluation metric. The results also separate properties that are often conflated: an imposed prior can remain inspectable without establishing physical fidelity, and retrieval can improve precision--recall ranking without improving thresholded F1. These distinctions provide a more precise basis for evaluating domain-informed model additions than aggregate accuracy alone.

The practical contribution of the research is a framework that can support fire behavior analysts by exposing two specific objects for review. The first is directional preferences introduced into attention and the second are the historical tiles used to correct a forecast. These objects allow an analyst to examine selected model assumptions and the relevance of retrieved examples alongside the predicted fire map. The use of modular additions also creates options for selecting a model according to predictive performance, computational resources, and the inspection functions required by a workflow. This study evaluates those capabilities retrospectively; it does not demonstrate that they improve evacuation decisions, crew deployment, or analyst performance. Their practical value should therefore be tested through supervised exercises and prospective shadow evaluations that measure error recognition, analyst workload, and decision quality. Operational use would additionally require timely inputs, validated probability estimates, and clear procedures for reviewing uncertain or poorly supported forecasts.

Implementation should begin with a retrospective pilot involving fire behavior analysts, model developers, and the organizations responsible for operational data. This phase should verify input availability at forecast issuance, data quality, regional performance, probability reliability, and the independence and relevance of retrieved examples. A subsequent shadow deployment could present forecasts alongside the prescribed directional bias, the environmental descriptors of retrieved tiles, and the magnitude of each retrieval correction, while retaining existing operational decision procedures. Analyst review should assess whether these displays help identify errors or unsupported forecasts and whether their interpretation adds excessive workload. Wider integration should depend on demonstrated benefit, documented failure handling, and monitoring of data changes, retrieval coverage, and performance across incidents. The system should remain a supervised analytical aid unless additional validation establishes suitability for a more consequential operational role.

This study is not without limitations. For example, its conclusions are bounded by its data, evaluation design, and the properties of the proposed components. Evaluation on one benchmark of daily, kilometer-scale fire observations does not establish transfer to finer-scale fire behavior or future incidents; additional datasets and event-disjoint or chronological evaluations are needed to assess those settings. The retrieval bank draws on the available training and validation distribution, so train-only-bank and independent-event evaluations are needed to distinguish transferable analog information from dependence on related samples. The supplementary analysis assesses, through five control experiments, to what extent the PFRAOC gain stems from the specific analogs selected. It finds that physics-matched tiles are statistically indistinguishable from random tiles in accuracy; however, they overlap the observed next-day fire mask significantly more than random tiles, supporting their role as inspection objects.  Because directional alignment is measured on a bias constructed from environmental inputs, it supports inspection of that prior but not a complete explanation of the forecast; output-level perturbation tests can examine how the prior affects predicted fire activity. That alignment is also checked against field-measurable regional wind, not the fire-induced local winds that can decouple from GRIDMET-scale products during plume-driven behavior; whether the audit remains useful in those regimes is an open operational question. Small predictive differences also require paired uncertainty estimates that distinguish training variability from dependence among test samples. Finally, aggregate calibration measures and retrospective visualizations do not establish operational benefit, which should be evaluated through probability diagnostics, supervised analyst studies, and prospective shadow trials.

% ============================================================

% ============================================================
\section{Conclusion}
\label{sec:conclusion}
% ============================================================
This study examined how explicit directional priors, retrieved fire examples, and fire-conditioned gating can be incorporated into next-day wildfire prediction models. The five backbone comparison shows that these components can be integrated into multiple architectures. This not only advances the wildfire domain, but could be modified for other fields. This can be supported by the fact that the MK-UNet and SwinUNETR are adapted from the medical domain, thus the proposed tools can be modified for other segmentation tasks. The predictive effect in this study is host-dependent. The full mechanism stack produced the strongest individual result on SwinUNETR, while the same additions modestly degraded CAFIM-CNN's classification accuracy. The attention bias exposes a prescribed environmental preference. Specifically, the wind bias aligns with the observed wind direction in roughly 98\% of test tiles. Retrieval provides identifiable examples associated with a learned correction, whose dependence on physically coherent retrieval is confirmed by a wrong-physics ablation. The central contribution of this research is a framework for examining selected domain-informed components together with their empirical benefits and limitations. This is illustrated by a mixed ensemble of mechanism-equipped and unmodified backbones that reaches the best classification result in the study (F1 $=$ 0.4292 $\pm$ 0.0016, AUC-PR $=$ 0.3790 $\pm$ 0.0025, confirmed across three seeds) while remaining fully auditable, since its output is an unweighted mean of its members' predictions. Further progress should connect these inspection capabilities to independently validated predictive behavior and measurable improvements in analyst interpretation.

% ============================================================
% BIBLIOGRAPHY
% ============================================================
\bibliographystyle{unsrt}
\bibliography{references}

\clearpage
\onecolumn
\section*{Supplementary Material}
\label{appendix}
\begin{figure*}[h]
\centering
\includegraphics[width=\textwidth]{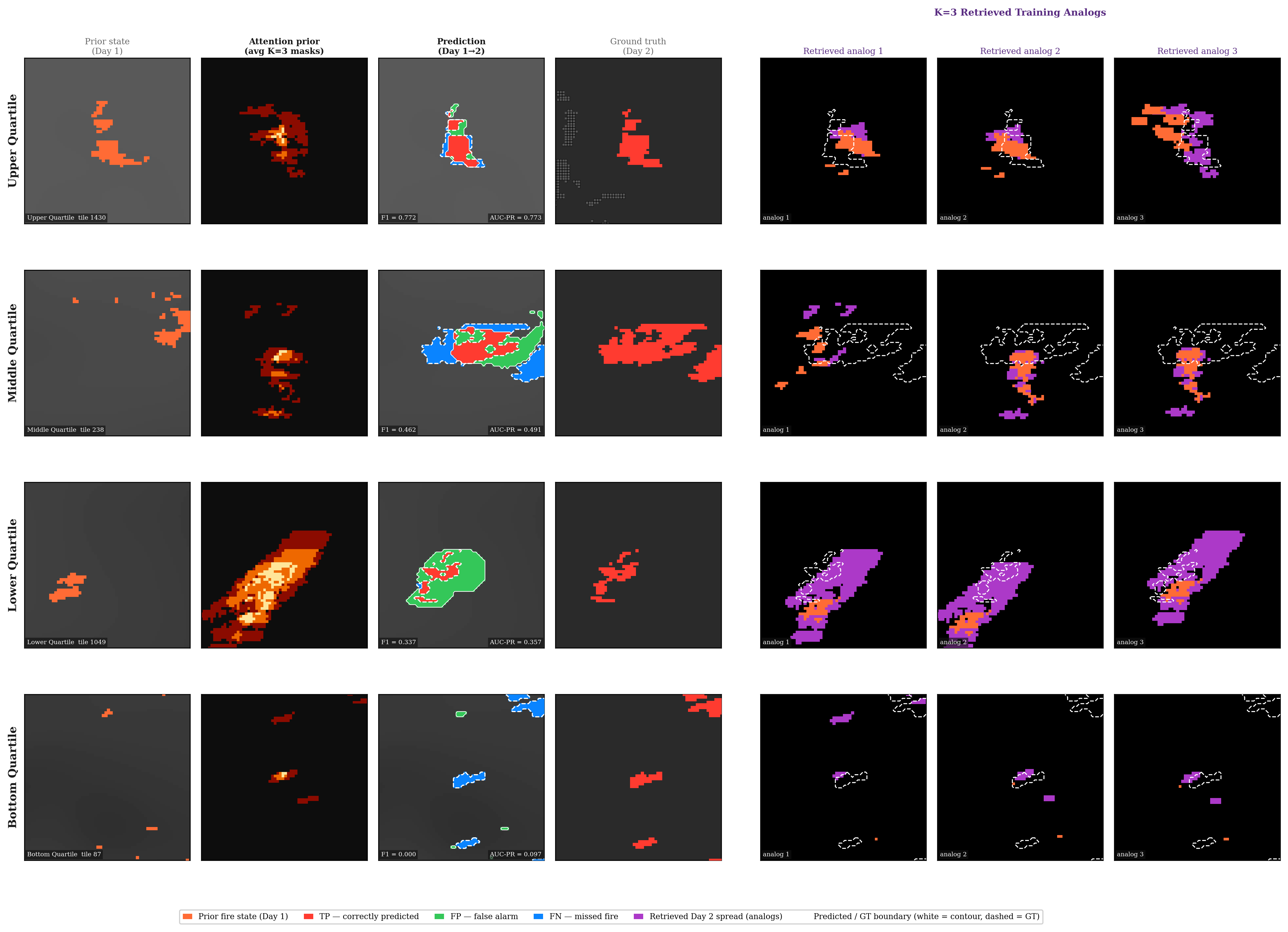}
\caption{Physics-keyed retrieval-augmented feature correction
walkthrough across the upper, middle, lower, and bottom F1 quartiles of
the test set. Beyond the prior state, prediction, and ground truth,
each row surfaces the $K{=}3$ retrieved historical analog fires used to compute the correction, with the query tile's own Day-2 boundary
(dashed) overlaid on each analog for direct visual comparison.}
\label{fig:vuln_rag}
\end{figure*}

Figure \ref{fig:vuln_rag} shows the upper, middle, lower, and bottom quartile of the SwinUNETR model with all three mechanisms. Similar to Figure \ref{fig:TwoMechMap} the false negatives lie within the perimeter of the fire across a majority of the panel. This figure adds the inspectable analogs. The retrieved fires are not geometric matches for the query, as fire geometry varies too widely for exact correspondence, but they give the practitioner concrete reference cases retrieved under comparable conditions. In the upper-quartile case, the retrieved analogs overlap the query's observed Day-2 extent closely. The fire is large and well established, which is precisely the regime in which the retrieval-quality analysis below finds physics matched analogs most informative. In the remaining quartiles the geometric overlap is limited. There, the analogs' value lies less in shape correspondence than in showing how fire under similar wind, dryness, and terrain conditions progressed. This is the type of comparison the physic descriptor is designed to support, and one the analyst can weigh alongside the model's own forecast.

\begin{figure*}[t]
\centering
\includegraphics[width=\textwidth]{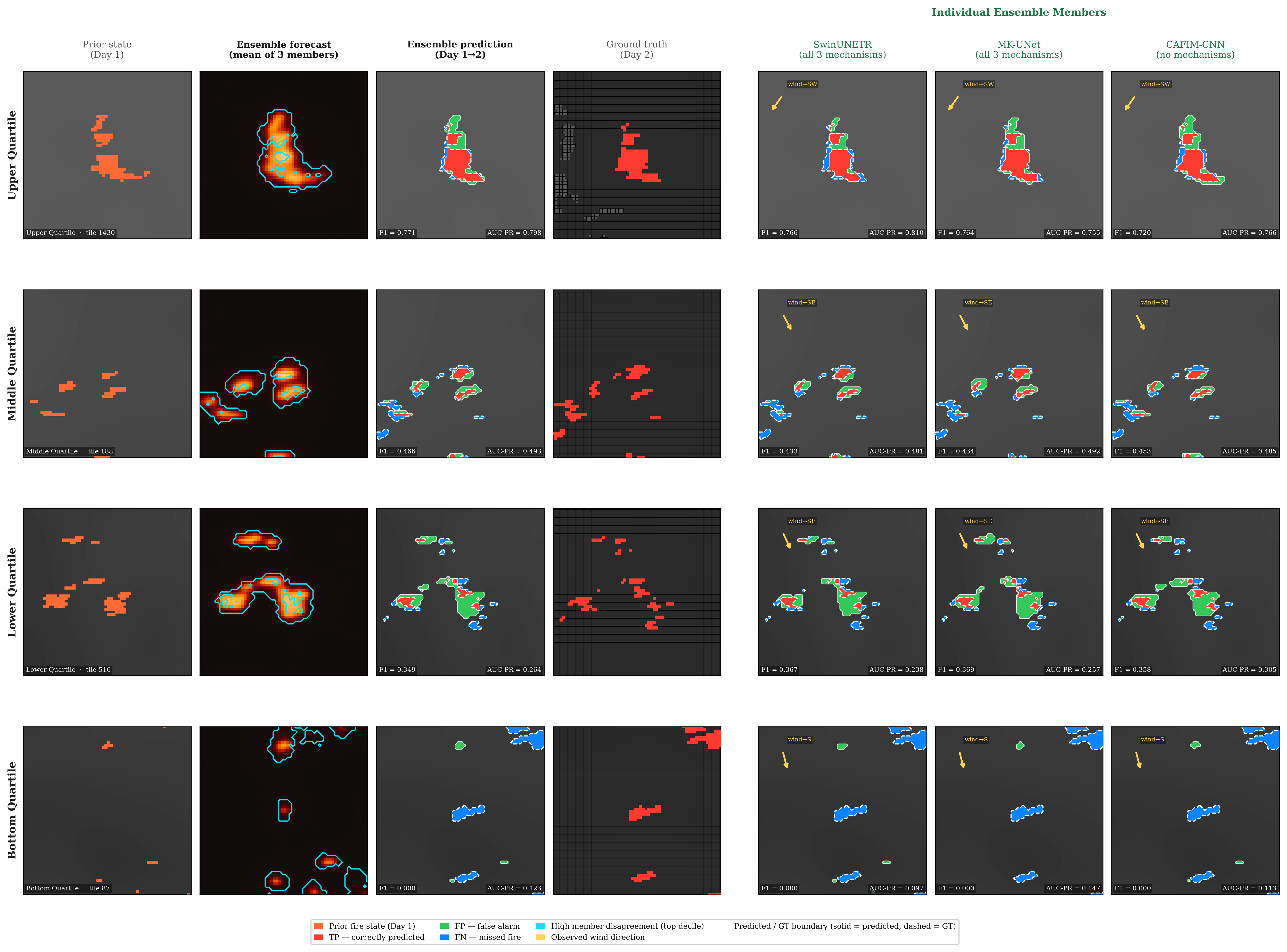}
\caption{Transparent decomposition of the mixed ensemble's forecast
into its three members (SwinUNETR and MK-UNet, both with all three
mechanisms; CAFIM-CNN, unmodified), across the upper, middle, lower,
and bottom F1 quartiles of the ensemble's own per-tile performance.
The Ensemble Forecast panel folds the members' pixel-wise disagreement
into a contour overlay rather than a separate panel; each member panel
additionally reports the observed wind direction so its predicted
spread can be checked against it.}
\label{fig:walkthrough_ensemble}
\end{figure*}

Figure ~\ref{fig:walkthrough_ensemble} demonstrates how the ensemble can be applied. Panel 1 shows the ongoing fire; panel~2 is the combined forecast, which sits between its three members on this tile. Its AUC-PR of 0.543 is slightly below the individual scores (SwinUNETR: 0.585; MK-UNet: 0.457; CAFIM-CNN: 0.559). The ensemble's advantage is not that it wins on every individual tile, but that it is more consistent across the full test set. Pooled over all test tiles, it exceeds every individual member (AUC-PR 0.381 versus 0.371/0.356/0.358) while remaining fully decomposable, so a forecaster can still open and inspect one model at a time. Panel~3 shows the per-pixel standard deviation across the three members, which concentrates along the fire perimeter. Panel~4 shows the three member forecasts side by side, each individually auditable. Each retrieval correction and attention bias can also be examined.

\newpage
Two complementary analyses characterize what the physical retrieval contributes. The first determines whether the correction's accuracy depends on which tiles are retrieved. Under matched correction capacity and training protocol, the research compares correct physics-matched retrieval against five content controls: (1) random-neighbor retrieval from the full bank, (2) a fixed dataset-mean mask, (3) a train-only bank, (4) random retrieval restricted to the same fire-containing candidate pools as the physics arm (5) a stricter subset with at least 5\% burning pixels. All five experiments are statistically indistinguishable from correct retrieval in test AUC-PR and F1, with overlapping bootstrap confidence intervals in every case. 

The second analysis determines whether the physical descriptors retrieve genuinely analogous fires. Using held-out test tiles as queries and restricting both selection rules to the same fire-containing candidate pool (bank fire fraction $>5\%$,
$n_{\mathrm{pool}}{=}644$), the top-$K{=}3$  physics matched tiles overlap the query's observed next-day fire mask modestly, but highly significantly better than random draws from the identical pool: mean IoU 0.086 vs.0.072(+21\% relative) over the 562 test queries with at least 1\% burning pixels (one-sided Wilcoxon signed-rank $p = 4\times10^{-12}$), and
0.112 vs.\ 0.098 at a stricter $\geq 5\%$ query floor ($n{=}81$,
$p{=}0.01$). Within the bank itself, where tiles from related incidents
are present, the corresponding gap is much larger (IoU 0.326 vs.\ 0.098);
we report this only as a within-bank upper bound, since it reflects the
related-incident dependence discussed in the main text rather than
generalizable analog quality.
\begin{table*}[t]
\centering
\small
\caption{Retrieval-content controls on the SwinUNETR host. All
configurations share the training method, threshold-selection protocol,
and evaluation tiles of the correct-retrieval result. All controls are statistically
indistinguishable from correct retrieval.}
\label{tab:supp_controls}
\setlength{\tabcolsep}{4pt}
\begin{tabular}{llccc}
\toprule
Retrieval content & Candidate pool & Seeds & F1 & AUC-PR \\
\midrule
Correct physics retrieval & full bank (16{,}856) & 3 & 0.4216 $\pm$ 0.0028 & 0.3673 $\pm$ 0.0033 [0.3534, 0.3871] \\
Random neighbor           & full bank            & 3 & 0.4237 $\pm$ 0.0004 & 0.3662 $\pm$ 0.0010 [0.3476, 0.3816] \\
Fixed mean mask           & ---                  & 3 & 0.4235 $\pm$ 0.0005 & 0.3658 $\pm$ 0.0006 [0.3476, 0.3816] \\
Train-only bank           & train split (14{,}979) & 3 & 0.4248 $\pm$ 0.0009 & 0.3679 $\pm$ 0.0017 [0.3507, 0.3842] \\
\midrule
Physics retrieval & fire-containing (88.7\%) & 1 & 0.4269 & 0.3709 [0.3535, 0.3870] \\
Random draw       & fire-containing (88.7\%) & 1 & 0.4239 & 0.3652 [0.3476, 0.3816] \\
Physics retrieval & fire frac $\geq 5\%$ (3.8\%) & 1 & 0.4239 & 0.3652 [0.3476, 0.3816] \\
Random draw       & fire frac $\geq 5\%$ (3.8\%) & 1 & 0.4239 & 0.3652 [0.3476, 0.3816] \\
\midrule
Physics retrieval & fire frac $\geq 5\%$ & 1 & 0.4247 & 0.3667 [0.3492, 0.3831] \\
Random draw       & fire frac $\geq 5\%$ & 1 & 0.4247 & 0.3667 [0.3492, 0.3831] \\
\bottomrule
\end{tabular}
\\[2pt]
\end{table*}

Taken together, the two analyses support the physical descriptor as the retrieval key. Moreover, the descriptor costs nothing in accuracy relative to any content control and yields analogs with measurably higher physical correspondence to the observed outcome, thus supporting their role as inspection objects. However, the current correction network, which consumes only the pixelwise mean of the retrieved masks, does not convert this retrieval-quality difference into an accuracy difference. Curating a denser and more behaviorally diverse bank would allow the retrieval content to contribute to accuracy. 

\end{document}